\documentclass[preprint,1p,11pt]{ISAS_IR}

\usepackage{abbreviations-paper}
\usepackage[utf8]{inputenc}
\usepackage{graphics}
\usepackage{epsfig} 
\usepackage[keeplastbox]{flushend}
\usepackage{amsmath} 
\usepackage{amssymb}  
\usepackage{bm}
\usepackage{subcaption}
\usepackage{textcomp}
\usepackage[linesnumbered,ruled,vlined]{algorithm2e}
\usepackage{url}
\usepackage{color} 
\usepackage[font=normalsize]{caption}
\usepackage{graphicx}
\usepackage{theorem}
\usepackage{booktabs}
\usepackage[english]{babel}
\usepackage{dsfont}
\usepackage{microtype}
\usepackage{xcolor}
\usepackage[hidelinks]{hyperref}
\usepackage{pifont}
\usepackage[cal=pxtx,scr=boondox]{mathalpha}
\usepackage{dutchcal}
\usepackage{float}
\usepackage{subfloat}
\usepackage{adjustbox}
\usepackage{dblfloatfix} 
\usepackage{mathtools}
\usepackage{tikz}

\newcommand{\bfX}{{\breve{\fX}}}
\newcommand{\bvx}{\breve{\vx}}
\newcommand{\bvp}{\breve{\vp}}
\newcommand{\bvq}{\breve{\vq}}
\newcommand{\bvv}{\breve{\vv}}
\newcommand{\bvb}{\breve{\vb}}
\newcommand{\bvt}{\breve{\vt}}
\newcommand{\Tg}{\text{g}}
\newcommand{\Ta}{\text{a}}
\newcommand{\Ts}{\text{s}}

\newcommand{\Tte}{\text{e}}
\newcommand{\Tw}{\text{w}}
\newcommand{\mW}{\mathbb{W}}
\newcommand{\mP}{\mathbb{P}}
\newcommand{\mE}{\mathbb{E}}
\newcommand{\mF}{\mathbb{F}}
\newcommand{\mG}{\mathbb{G}}
\newcommand{\mM}{\mathbb{M}}
\newcommand{\mS}{\mathbb{S}}
\newcommand{\srv}{\mathscr{v}}

\newcommand{\calR}{\mathcal{R}}
\newcommand{\calJ}{\mathcal{J}}
\newcommand{\calD}{\mathcal{D}}
\newcommand{\cl}{\mathbcal{l}}
\newcommand{\cw}{\mathbcal{w}}
\newcommand{\balpha}{\bm{\alpha}}
\newcommand{\bbeta}{\bm{\beta}}
\newcommand{\bgamma}{\bm{\gamma}}
\newcommand{\bnu}{\bm{\nu}}
\newcommand{\xmark}{\ding{55}}
\newcommand{\listar}{{\text{LiLi-OM}^\star}}

\begin{document}
	
	\begin{tikzpicture}[overlay, remember picture]
	\path (current page.north east) ++(-3.2,-0.4) node[below left] {
		Please cite the paper as: Kailai Li, Meng Li and Uwe D. Hanebeck,
	};
	\end{tikzpicture}
	\begin{tikzpicture}[overlay, remember picture]
	\path (current page.north east) ++(-2,-0.8) node[below left] {
		``Towards High-Performance Solid-State-LiDAR-Inertial Odometry and Mapping,''
	};
	\end{tikzpicture}
	\begin{tikzpicture}[overlay, remember picture]
	\path (current page.north east) ++(-2,-1.2) node[below left] {
 in IEEE Robotics and Automation Letters, vol. 6, no. 3, pp. 5167-5174, July 2021
	};
	\end{tikzpicture}
	
	\begin{frontmatter}
		
		\title{Towards High-Performance Solid-State-LiDAR-Inertial Odometry and Mapping}		
		
		\author{{Kailai~Li}, {Meng~Li}, and {Uwe~D.~Hanebeck}}
		
		\address{Intelligent Sensor-Actuator-Systems Laboratory (ISAS)\\
			Institute for Anthropomatics and Robotics\\
			Karlsruhe Institute of Technology (KIT), Germany\\
			emails: kailai.li@kit.edu, uyjjl@student.kit.edu, uwe.hanebeck@kit.edu}
		
		\begin{abstract}
			We present a novel tightly-coupled LiDAR-inertial odometry and mapping scheme for both solid-state and mechanical LiDARs. As frontend, a feature-based lightweight LiDAR odometry provides fast motion estimates for adaptive keyframe selection. As backend, a hierarchical keyframe-based sliding window optimization is performed through marginalization for directly fusing IMU and LiDAR measurements. For the Livox Horizon, a newly released solid-state LiDAR, a novel feature extraction method is proposed to handle its irregular scan pattern during preprocessing. LiLi-OM (Livox LiDAR-inertial odometry and mapping) is real-time capable and achieves superior accuracy over state-of-the-art systems for both LiDAR types on public data sets of mechanical LiDARs and in experiments using the Livox Horizon. Source code and recorded experimental data sets are available at https://github.com/KIT-ISAS/lili-om.
		\end{abstract}
		
	\end{frontmatter}
	
	
\section{Introduction} \label{sec:introduction}

Estimating six-DoF egomotion plays a fundamental role in a wealth of applications ranging from robot navigation and inspection to virtual/augmented reality, see~\cite{engel12vicomor,Fusion19_Bultmann,voxgraph2020,KahlerPVM16}. With the booming of autonomous driving, light detection and ranging (LiDAR) sensors have gained tremendous popularity~\cite{wen2020urbanloco,kitti}. Compared to visual sensors, 3D LiDARs provide lighting-invariant and accurate perception of the surroundings with long detection range and high robustness. Thus, they are broadly deployed to mobile agents for odometry and mapping.

Essentially, LiDAR-based odometry requires computing six-DoF egomotion given consecutive frames of point clouds. This can be performed via scan-matching algorithms, e.g., the iterative closest point (ICP)~\cite{Pomerleau12comp} method. In typical mobile perception scenarios, 3D LiDARs output a large streaming volume of raw scans in the form of unorganized point clouds. Meanwhile, real-time processing (e.g., at $10$ Hz) is required given limited computing resources. To improve the performance of LiDAR odometry, much attention has been dedicated to scan-matching with points representing local geometric features~\cite{segal2009generalized,chen1992object}. In~\cite{zhang2014loam}, feature points are extracted from object edges and planes for scan-matching using point-to-edge and point-to-plane metrics, which enable accurate LiDAR odometry in real time. Based thereon, in~\cite{legoloam2018shan}, an image-based approach~\cite{bogoslavskyi2016fast} was further applied to preprocessing. A two-stage optimization scheme was tailored for ground vehicles to enable light-weight, robust LiDAR odometry and mapping.

\begin{figure} 
	\centering
	\includegraphics[width=0.95\textwidth]{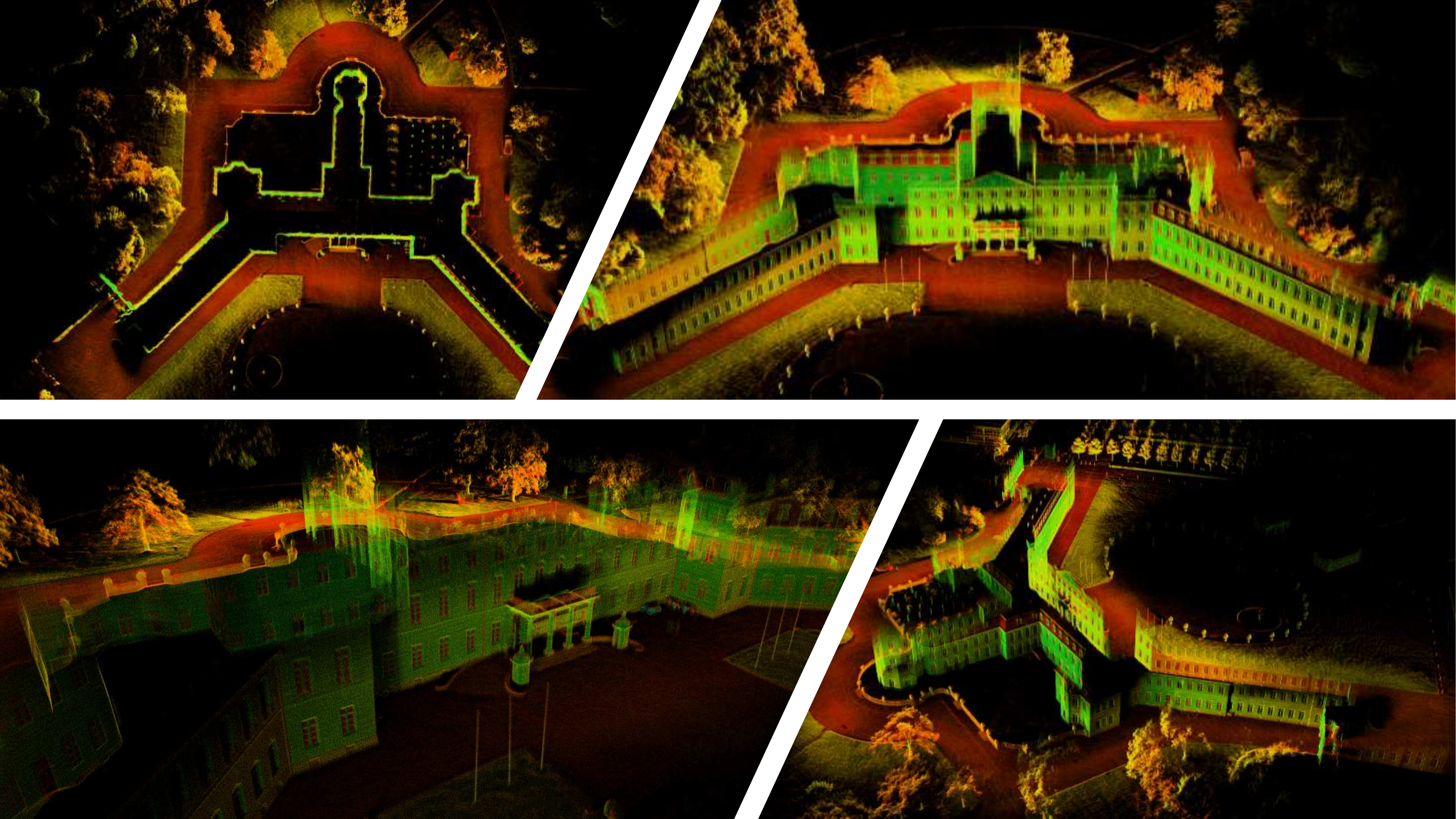} 
	\caption{3D map of Schloss Karlsruhe from LiLi-OM.} 
	\label{fig:front}	
\end{figure}

Typical LiDAR scan rates are relatively low and the perceived point clouds are in principle distorted due to sensor egomotion. Thus, performing LiDAR-only odometry is prone to deterioration under fast motion or in complex scenes. Inertial sensors, however, measure instant motion at a much higher frequency and can bridge the gap between consecutive LiDAR frames, improving the robustness and accuracy of LiDAR-based egomotion estimation. In~\cite{zhang2014loam,legoloam2018shan}, an inertial measurement unit (IMU) was used to de-skew point clouds and initialize motion estimates for scan-matching-based LiDAR odometry. Though decoupled or loosely-coupled LiDAR-inertial fusion schemes are appealing for runtime and deploying classic recursive filters (e.g., the EKF), they may cause information loss and inaccurate estimates~\cite{tang2015lidar}.

Thus, there has been a growing focus on tightly-coupled LiDAR-inertial odometry, where  point cloud and IMU measurements are fused in a joint optimization or filtering framework. Pre-integrated IMU readings are often employed for de-skewing the LiDAR scan per frame~\cite{le2020in2laama}. In~\cite{ye2019tightly}, an optimization-based approach was proposed for LiDAR-inertial odometry using a maximum a posteriori (MAP) formulation incorporating both LiDAR and IMU residuals in a sliding window fashion. An additional method using rotational constraints was proposed to refine the final pose and map, which delivers similar or better tracking accuracy as~\cite{zhang2014loam}. However, real-time processing is hard to achieve in practice as sensor readings of every frame are exploited. 

To improve the runtime efficiency of LiDAR-inertial odometry, an iterated error-state Kalman filter was introduced in~\cite{lins2020} based on a robocentric formulation. It runs in real time and shows superior tracking accuracy over existing LiDAR-only odometry systems. In~\cite{liosam2020shan}, a tightly-coupled LiDAR-inertial odometry system was proposed based on the incremental smoothing and mapping framework iSAM2~\cite{kaess2012isam2}. However, the system relies heavily on~\cite{zhang2014loam} to produce LiDAR odometry factors for further constraining the pre-integrated IMU states in the factor graph formulation. Unlike direct fusion of IMU and LiDAR measurements within a unified scheme, this can result in loss of constraint information posed by landmarks. Also, correlations between LiDAR and IMU measurements might be largely discarded. To guarantee high odometry accuracy, the system requires nine-axis IMU readings of high frequency ($500$ Hz as used in~\cite{liosam2020shan}) to de-skew the point cloud and initialize LiDAR odometry. Fusion with additional sensor modalities (e.g., GPS signals) is often needed at certain spots to achieve precise localization. 

More importantly, conventional LiDARs rely on mechanical spinning mechanisms to enable a $360$-degree FoV, while the vertical resolution is fairly limited. Though some products with high vertical resolution emerged recently (e.g., from Hesai\footnote{{\tt https://www.hesaitech.com/en/Pandar128}} and Velodyne\footnote{\tt https://velodynelidar.com/products/alpha-prime}), their very high prices prohibit mass market supply for robotics industry and research.

Very recently, \emph{solid-state LiDAR}s have hit the consumer market with much better affordability based on various working principles. Existing types often have non-repetitive and irregular scan patterns with small FoVs to reach more uniform and higher resolution. So far, solid-state-LiDAR-based odometry has not been well investigated. In~\cite{livoloam2019}, the LiDAR odometry system in~\cite{zhang2014loam} was adapted to Livox Mid-40\footnote{\tt https://www.livoxtech.com/mid-40-and-mid-100}, a solid-state LiDAR with a circular FoV of $38.4^\circ$. Compared with its baseline~\cite{zhang2014loam}, it employs similar feature-based scan-matching and delivers comparable tracking accuracy with improved runtime via parallelization. To the best of the authors' knowledge, no published research on tightly-coupled solid-state-LiDAR-inertial fusion exists to date.

In this paper, we provide a specific study on solid-state-LiDAR-inertial odometry and mapping. Instead of Livox Mid-40, we choose Livox Horizon\footnote{\tt https://www.livoxtech.com/de/horizon} (released in Q1, 2020), which is designed for vehicular perception with an FoV of $81.7^\circ\times25.1^\circ$. Scanning at $10$ Hz, it reaches a similar but more uniform FoV coverage compared with typical $64$-line mechanical LiDARs. Moreover, it is substantially cheaper than most existing 3D LiDARs of comparable performance.

Common feature extraction methods for spinning LiDARs are not applicable for solid-state ones due to their irregular and non-repetitive scan patterns. For Livox Mid-40, scanning of a single laser head is specially regulated to form a circular coverage. Hence, the approach in~\cite{livoloam2019} traverses along the incident and deflection angle to choose point candidates, at which the local smoothness of the scan line is computed for feature extraction as in~\cite{zhang2014loam}. This, however, cannot be applied to Livox Horizon as it sweeps in a rather unregulated manner. Such a pattern is more generic for reaching broader FoVs of uniform coverage, which can potentially become common for future types of solid-state LiDARs. But the limited FoVs can still thwart odometry performance in some circumstances, especially under fast motion or with insufficient features.

\subsection*{Contributions}
Considering the state of the art above, we propose a novel tightly-coupled LiDAR-inertial odometry and mapping scheme with a specific variant for solid-state LiDARs (pipeline given in Sec.\,\ref{sec:pipe}). A novel feature extraction approach is tailored to the irregular scan pattern of Livox Horizon (Sec.\,\ref{sec:ptMetric}). To directly fuse LiDAR and IMU measurements in a unified manner, a hierarchical keyframe-based scheme is proposed using sliding window optimization (Sec.\,\ref{sec:keyframeOpt}). The proposed system is generically applicable for both conventional and the deployed solid-state LiDAR. It runs in real time and delivers superior odometry accuracy over existing systems (Sec.\,\ref{sec:eva}). We release the proposed system with open-source code and new solid-state-LiDAR-inertial data sets recorded by Livox Horizon and Xsens MTi-670. Thanks to the low hardware costs and real-time performance on portable platforms, our system provides a cost-effective solution for mobile perception in various scenarios.

\section{System Pipeline} \label{sec:pipe}

\begin{figure}[t]
	\centering
	\includegraphics[width=0.95\textwidth]{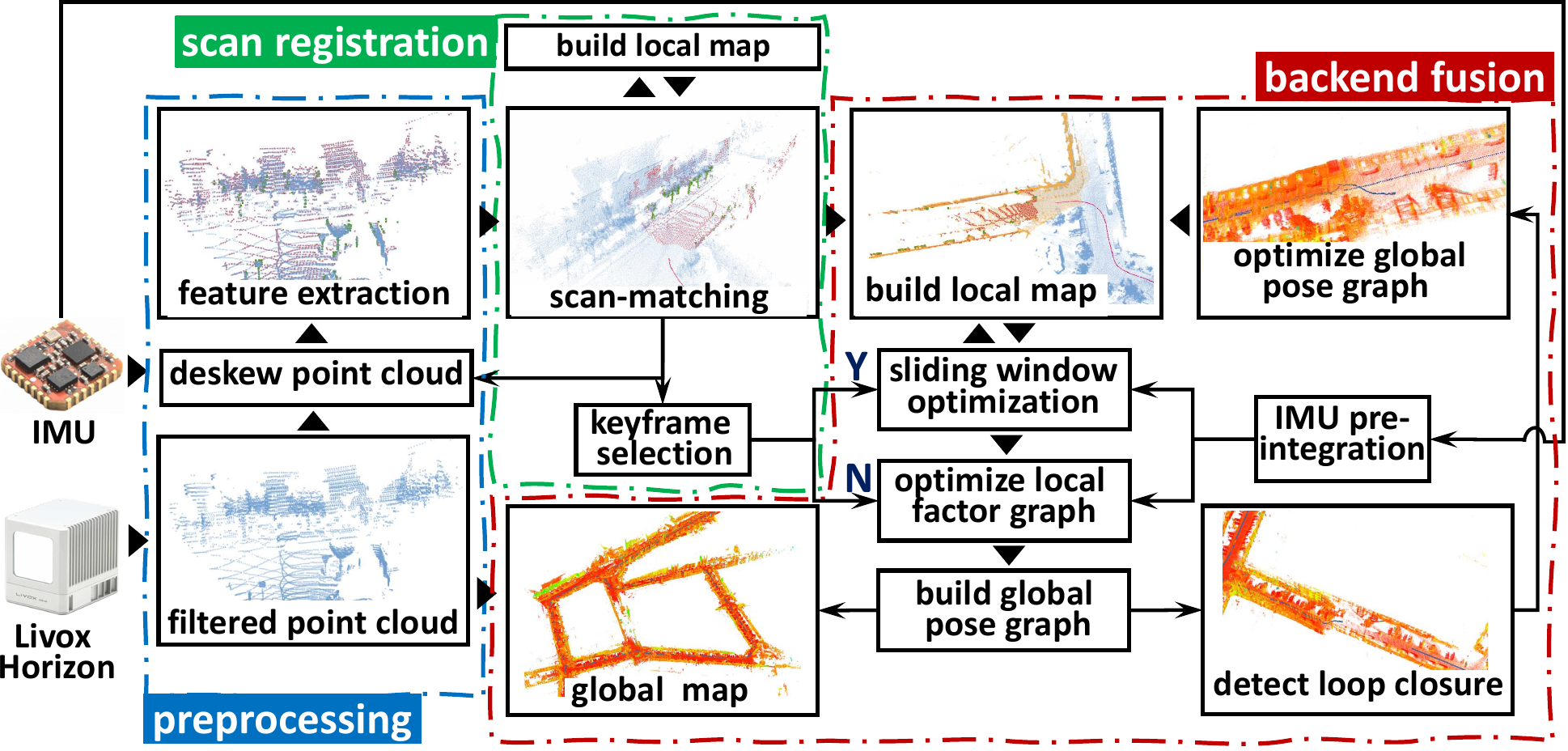} 
	\caption{System pipeline}	
	\label{fig:flow}	
\end{figure}

The proposed LiDAR-inertial odometry and mapping scheme is shown in Fig.\,\ref{fig:flow}. The 3D LiDAR (e.g., Livox Horizon) streams out point clouds at a typical frequency of $10$ Hz and is synchronized with a six-axis IMU providing gyroscope and accelerometer readings at higher frequency (e.g., $200$ Hz for Xsens MTi-670\footnote{\tt https://www.xsens.com/products/mti-600-series}). We want to estimate the six-DoF egomotion of the LiDAR frame and obtain a globally consistent map simultaneously. The raw point clouds from LiDAR scan are first downsampled and rotationally de-skewed using gyroscope data. Then, feature points representing planes and edges are extracted (Sec.\,\ref{subsec:feature}). Given the preprocessed scans, a light-weight scan-matching-based registration module runs in a frame-to-model manner for fast motion estimation with point-to-edge and point-to-plane metrics being exploited (Sec.\,\ref{subsec:lidarMetric}). The obtained egomotion estimates are further used to de-skew the translational distortion of the current sweep as well as select keyframes adaptively to scene transitions. Parallel to preprocessing and LiDAR odometry, LiDAR and IMU measurements are fused at the backend in a unified manner via the proposed keyframe-based sliding window optimization (Sec.\,\ref{sec:keyframeOpt}). 

The fusion window usually covers several (e.g., three) keyframes. As the window slides, keyframe states (denoted with $\breve{ }$ on top as follows) are optimized in the current window
\begin{equation} \label{eq:keyframe}
\breve{\vx}=\bbmat\,\bvt^\top,\bvv^\top,\bvq^\top,\bvb^\top\,\ebmat^\top\in\R^3\times\R^3\times\Sbb^3\times\R^6\subset\R^{16}\,.
\end{equation}
Here, $\bvt\in\R^3$ and $\bvq\in\Sbb^3$ denote the keyframe position and orientation (represented by unit quaternions), respectively. $\bvv\in\R^3$ is the velocity and $\bvb=[\,\bvb_\Ta^\top,\bvb_\Tg^\top\,]^\top\in\R^6$ denotes the term incorporating the IMU bias of the accelerometer (subscript `a') and the gyroscope (subscript `g'). After being slid over, the two oldest optimized keyframes are used as constraints to optimize the in-between regular-frame poses via factor graph optimization. Preintegrated inertial measurements are hereby exploited for pose initialization. A global pose graph is maintained to incorporate all poses of LiDAR frames. Loop closure is checked in a keyframe basis using ICP, and when necessary, a global graph optimization is invoked to guarantee the reconstructed map to be globally consistent.

\begin{figure} 
	\centering
	\begin{tabular}{cc}
		\adjustbox{trim={.01\width} {.01\height} {.1\width} {.01\height},clip}{\includegraphics[width=0.45\textwidth]{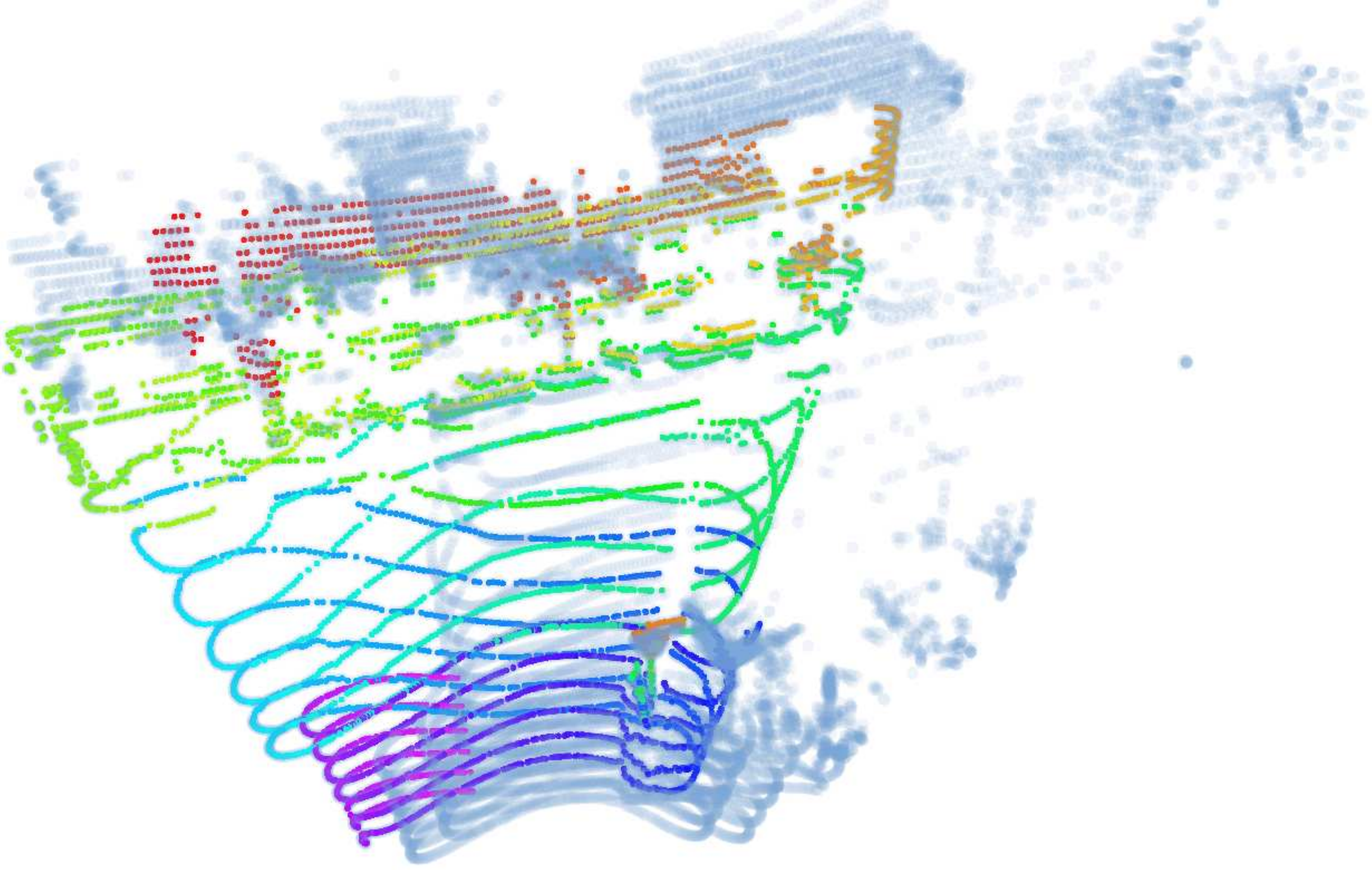}} &		
		\adjustbox{trim={.01\width} {.01\height} {.1\width} {.01\height},clip}{\includegraphics[width=0.45\textwidth]{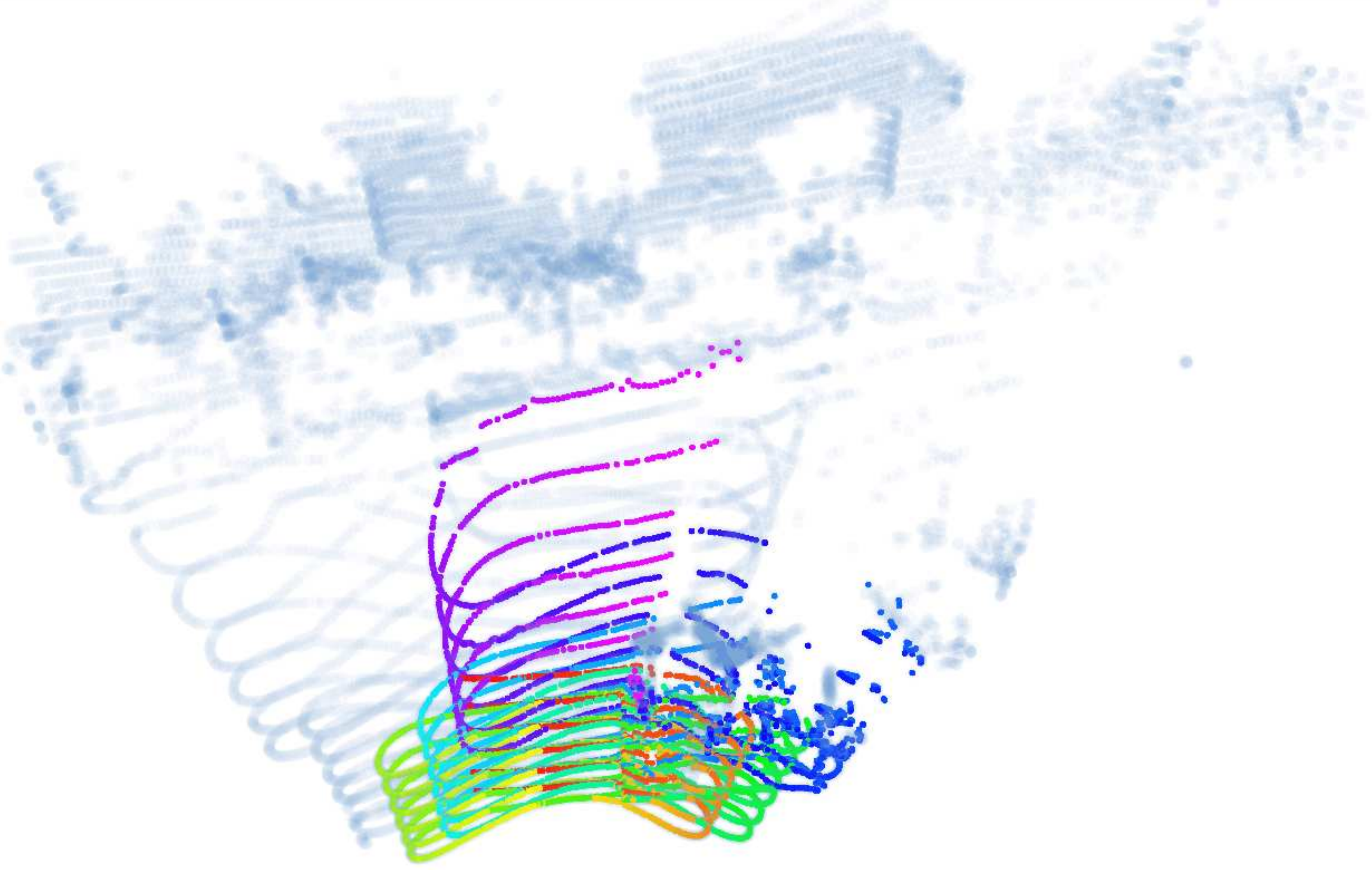}} \\	
		(A) $0$ - $25$ ms & (B) $25$ - $50$ ms\\
		\adjustbox{trim={.01\width} {.01\height} {.1\width} {.01\height},clip}{\includegraphics[width=0.45\textwidth]{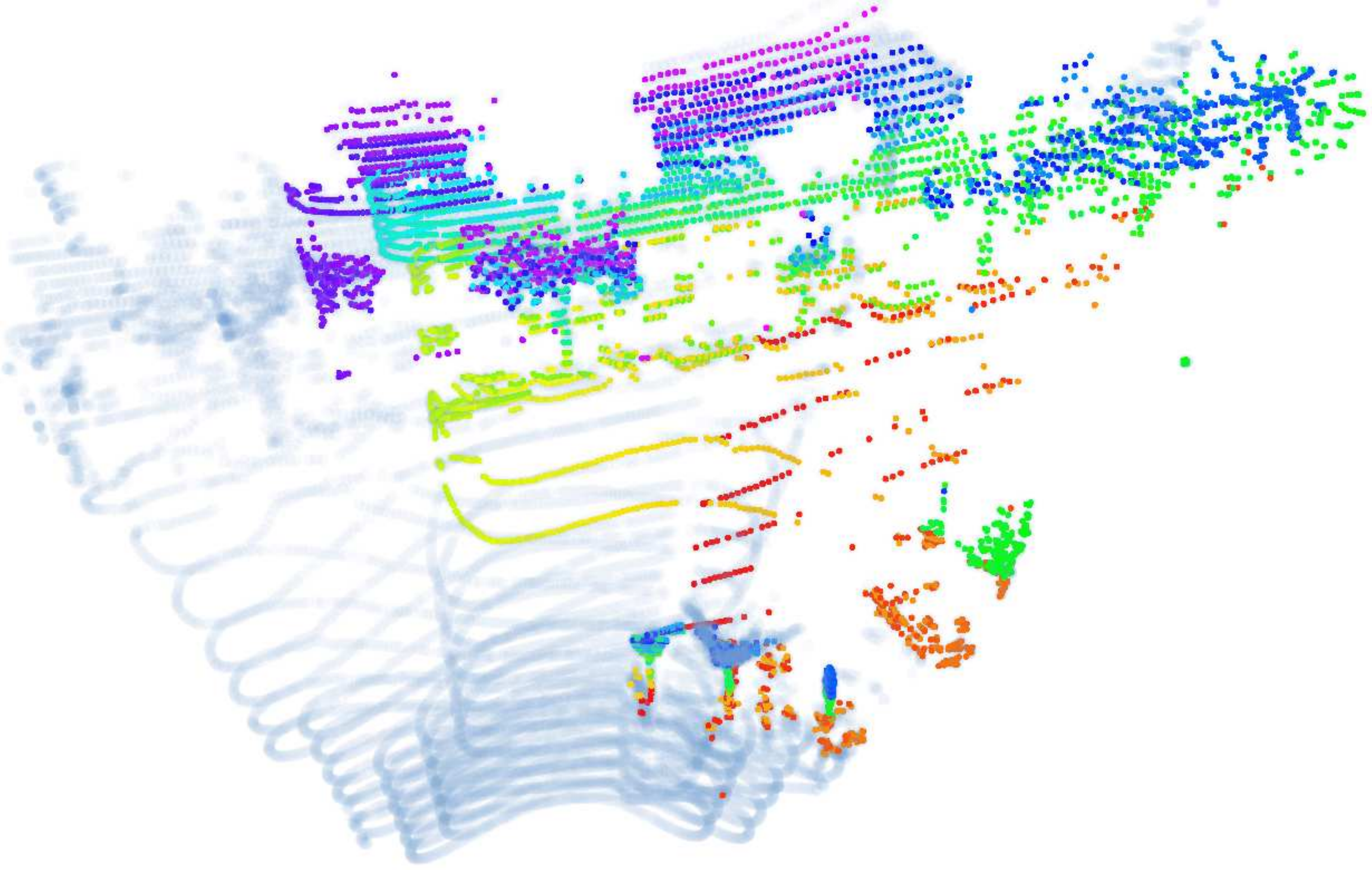}} &	
		\adjustbox{trim={.01\width} {.01\height} {.1\width} {.01\height},clip}{\includegraphics[width=0.45\textwidth]{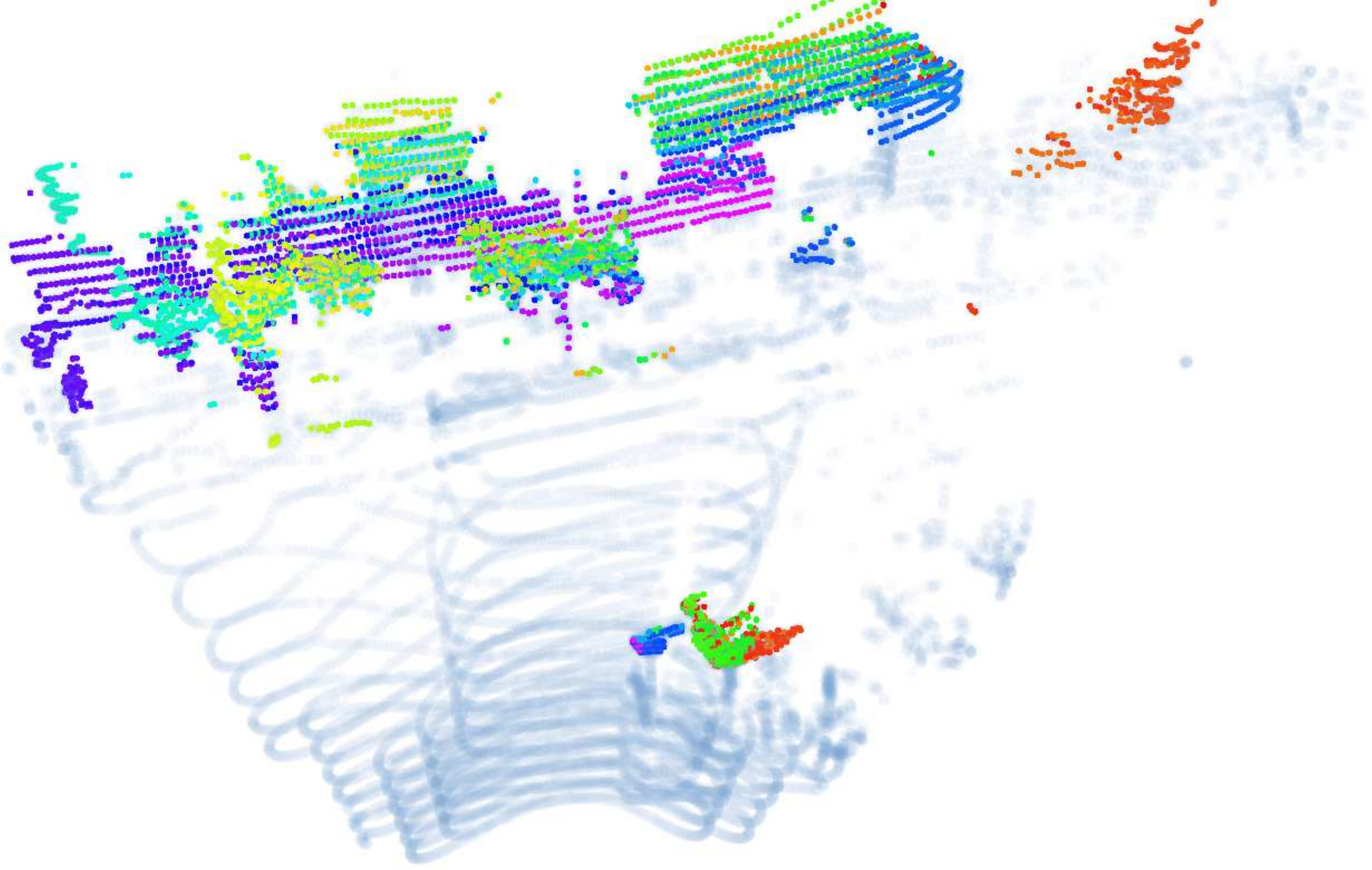}} \\		
		(C) $50$ - $75$ ms & (D) $75$ - $100$ ms		
	\end{tabular}
	\caption{Scan pattern of Livox Horizon with real-world scene. Points are rendered according to time stamps with rainbow color scale. Blue blobs depict the full sweep.}	
	\label{fig:scanPattern}	
\end{figure}
	
\section{Feature-Based Solid-State LiDAR Scan-Matching} \label{sec:ptMetric}
\subsection{Feature extraction for irregular scan pattern} \label{subsec:feature}	
Existing feature extraction methods for 3D LiDARs are not well applicable for Livox Horizon~\cite{legoloam2018shan,livoloam2019}. Fig.\,\ref{fig:scanPattern} illustrates its scan pattern (at $10$ Hz) by dividing one LiDAR sweep into four stages. The Livox Horizon is instrumented with a multi-laser sensing module with an array of six vertically-aligned laser diodes sweeping back and forth through the prisms non-repetitively. Consequently, the six point readings obtained from the multi-laser transceiver are vertically aligned and perceived simultaneously (rendered as the same color). The scan shows an irregular ``brushing'' pattern covering the $81.7^\circ\times25.1^\circ$ FoV uniformly. Over the integration time of $100$\,ms per frame, the angular resolution reaches $0.2^\circ$ to $0.4^\circ$ horizontally and vertically with a similar FoV coverage to $64$-line spinning LiDARs. Compared with Mid-40 that regulates its single laser head for circular FoV coverage, Horizon has a more irregular scan pattern due to its rather unregulated sweeping motion.	
\begin{figure}[t]
	\centering
	\includegraphics[width=0.95\textwidth]{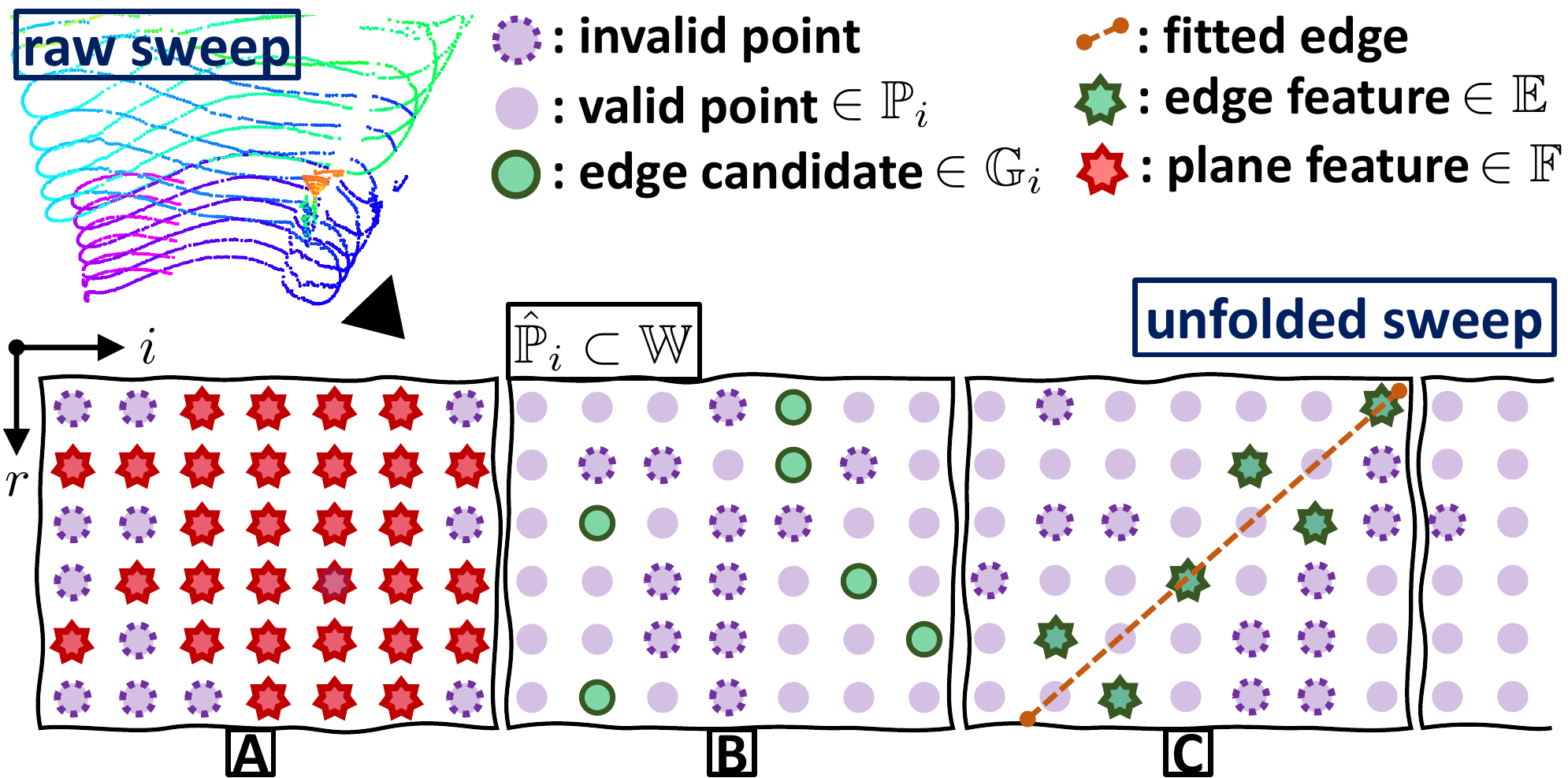} 
	\caption{Illustration of proposed feature extraction method.}
	\label{fig:extractionMethod}
\end{figure}

\LinesNumbered
\begin{algorithm}[t]
	\caption{\strut Feature Extraction for Livox Horizon} \label{alg:extraction}
	\KwIn{single sweep $\mW$}
	\KwOut{edge feature set $\mE$, plane feature set $\mF$}
	$\mE\gets\emptyset$,\,$\mF\gets\emptyset$\,\;
	$\{\hat{\mP}_i=\{\fX_r\}_{r=1}^6\,\vert\,\fX_r\in\R^{3\times7}\}_{i=1}^\tau\gets\texttt{split}\,(\mW)$\,\;
	\For{$i \gets 1$ \KwTo $\tau$}{
		$\mP_i\gets\texttt{getValidPoints}\,(\hat{\mP}_i)$\,\;
		$\Sigma_i\gets\texttt{computeCovariance}\,(\mP_i)$\,\;
		$\{(\lambda_1,\lambda_2,\lambda_3)\}\gets\texttt{eig}\,(\Sigma_i)$\tcp*{$\lambda_1\leq\lambda_2\leq\lambda_3$}
		\tcc{check plane feature}
		\eIf{$\lambda_1/\lambda_2<0.3$}
		{
			$\mF\gets\mF\cup \mP_i$\,\;
		}
		{
			\tcc{check edge feature}
			$\mG_i\gets\emptyset$\,\;
			\For{$r \gets 1$ \KwTo $6$}{
				$\vx_r\gets\texttt{getCurv}\,(\fX_r)$\,\;
				$\mG_i\gets\mG_i\cup\fX_r$\,\;
			}
			$\Gamma_i\gets\texttt{computeCovariance}\,(\mG_i)$\,\;
			$\{(\lambda_1,\lambda_2,\lambda_3)\}\gets\texttt{eig}\,(\Gamma_i)$\,\;
			\If{$\lambda_2/\lambda_3<0.25$}{
				$\mE\gets\mE\cup \mG_i$\,\;
			}
		}
	}
\end{algorithm}

To extract plane and edge feature points for Livox Horizon, we propose a new two-stage approach in Alg.\,\ref{alg:extraction} (illustrated in Fig.\,\ref{fig:extractionMethod}). Given a frame of raw scan, we unfold and split the sweep $\mW$ in its time domain, where $6\times7$-point patches are assigned one after another without overlapping (Alg.\,\ref{alg:extraction}, line 2). In each raw patch $\hat{\mP}_i$, inconcrete point readings are first removed (Alg.\,\ref{alg:extraction}, line 4). For the valid points $\mP_i$, we perform eigendecomposition of the covariance of their 3D coordinates (Alg.\,\ref{alg:extraction}, line 5-6). If the second largest eigenvalue is substantially larger than the smallest one ($\lambda_1/\lambda_2<0.3$), then all points in the patch are extracted as plane features  (Fig.\,\ref{fig:extractionMethod}-A and Alg.\,\ref{alg:extraction}, line 7-8). For a non-plane patch, we search the point with largest curvature on each scan line and perform eigendecomposition for the six points (Alg.\,\ref{alg:extraction}, line 9-15). If the largest eigenvalue is substantially larger than the second largest one ($\lambda_2/\lambda_3<0.25$), then the points form a line and they are extracted as edge features (Fig.\,\ref{fig:extractionMethod}-C and Alg.\,\ref{alg:extraction}, line 16-17). Otherwise, no feature is extracted from the current patch (Fig.\,\ref{fig:extractionMethod}-B).

\begin{figure}[t]
	\centering
	\begin{tabular}{c}
		\includegraphics[width=0.95\textwidth]{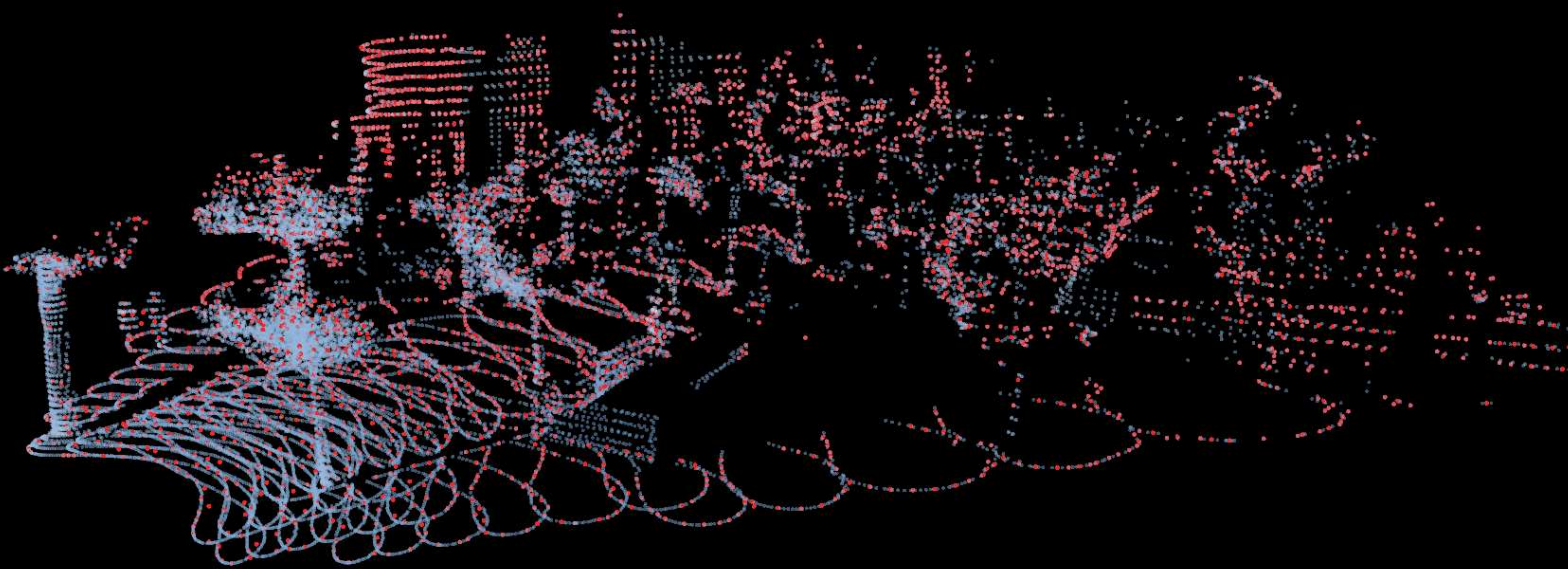} \\
		(A) extracted plane features (red)\\
		\includegraphics[width=0.95\textwidth]{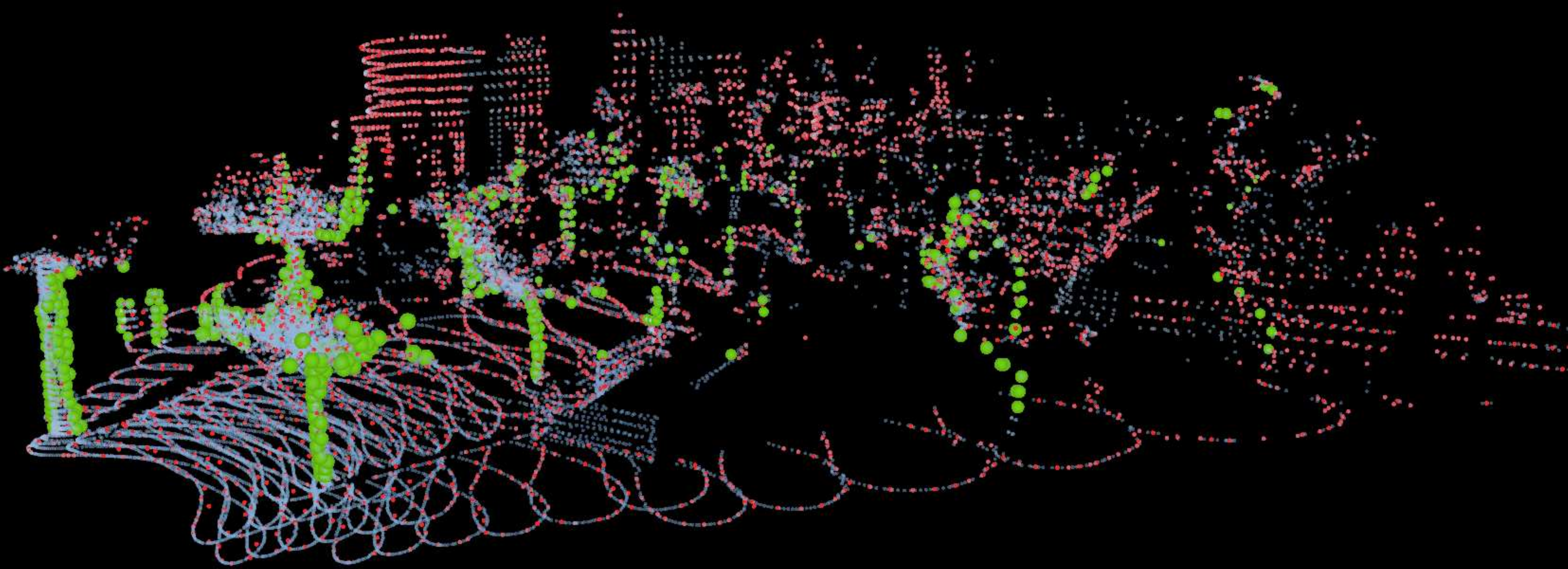} \\
		(B) extracted edge (green) and plane (red) features 
	\end{tabular}
	\caption{Feature extraction for point cloud (blue) from Horizon.}
	\label{fig:featureExtraction}
\end{figure}

We show an example of extracted features in one frame of Livox Horizon scan in Fig.\,\ref{fig:featureExtraction}. Note that the proposed feature extraction algorithm is purely performed in the time domain of every sweep given time stamps of the perceived point array. This is radically different from the approach in~\cite{livoloam2019} that computes local smoothness of point candidates selected through spatial retrieval. Also for traditional spinning LiDARs, common point cloud segmentation or feature extraction methods~\cite{zhang2014loam,legoloam2018shan,squeezSeg} are performed in (transformed) spatial domains, e.g., extracting features w.r.t. the horizontal scan angles. For Livox Horizon, the number of extracted plane features are usually much more than edge features due to its uniform scan coverage (statistics are given in Sec.\,\ref{subsec:runtime}). We further associate each feature point with its corresponding edge's direction vector or plane's normal vector to represent local geometries. It will be further exploited for weighting the LiDAR residual term in the backend fusion module. 

\subsection{Point-to-edge and point-to-plane metric} \label{subsec:lidarMetric}
Given the extracted edge and plane features, both the frontend registration module and the sliding window optimization for backend fusion in Fig.\,\ref{fig:flow} exploit the scan-matching formulation to incorporate LiDAR measurements for egomotion estimation. Following metrics are applied.

\subsubsection*{Point-to-edge metric} 
Suppose an edge feature $\vp^\cl\in\mE$ is extracted from sweep $\mW$ w.r.t. the LiDAR frame $\cl$ and its associated unit vector $\bnu_\Tte$ indicating the edge direction is available. We first search its nearest five edge points in the corresponding local feature map $\mM^\cw$ and compute their coordinates' mean value $\bar{\ve}^\cw$ and covariance matrix (w.r.t. the global frame $\cw$). Based thereon, an eigendecomposition is performed. If the largest eigenvalue is significantly larger than the rest, the five points in $\mM^\cw$ form a line with its direction vector $\vn_\Tte$ being the eigenvector corresponding to the largest eigenvalue. We then take two points $\acute{\ve}^\cw=\bar{\ve}^\cw+\delta\vn_\Tte$ and $\grave{\ve}^\cw=\bar{\ve}^\cw-\delta\vn_\Tte$ on the fitted line and exploit a point-to-edge metric of the following form
\begin{equation*}
\calD_\Tte(\vx^\cw,\vp^\cl,\mM^\cw)=\frac{\Vert(\vp^\cw-\acute{\ve}^\cw)\times(\vp^\cw-\grave{\ve}^\cw)\Vert}{\Vert\acute{\ve}^\cw-\grave{\ve}^\cw\Vert}\,.
\end{equation*}
Here, $\vp^\cw=\fR(\vq)\,\vp^\cl+\vt$ denotes the scan point w.r.t. the global frame given current LiDAR pose $\vx^\top=[\,\vq^\top,\vt^\top\,]^\top$. $\fR(\vq)$ is the rotation matrix given by $\vq$. Typically, $\delta=0.1$\,.

\subsubsection*{Point-to-plane metric}
If the feature point $\vp^\cl$ indicates a plane (with a normal $\bnu_\Ts$), we also transform it into its world coordinates $\vp^\cw$ and find its nearest five plane feature points in the corresponding local feature map w.r.t. the global frame, namely $\mS^\cw=\{\vs^\cw_j\}\subset\mM^\cw$, with $j=1,...,5$. Similar to the implementation for\cite{zhang2014loam}, we solve an overdetermined linear equation $\fA_\Ts\vu_\Ts=\vc$ via QR decomposition for plane fitting, with $\fA_\Ts=[\,\vs^\cw_1,...,\vs^\cw_5\,]\in\R^{5\times3}$ and $\vc=[\,-1,...,-1\,]\in\R^5$. We normalize the fitted normal vector as $\vn_\Ts=\vu_\Ts/\norm{\vu_\Ts}$ and the point-to-plane metric can be established as

\begin{equation*}
\calD_\Ts(\vx^\cw,\vp^\cl,\mM^\cw)=\vert\vu_\Ts^\top\vp^\cw+1\vert/\norm{\vu_\Ts}=\big\vert\vn_\Ts^\top\vp^\cw+1/\norm{\vu_\Ts}\big\vert\,.
\end{equation*}
Here, we also have $\vp^\cw=\fR(\vq)\,\vp^\cl+\vt$.

\subsection{Metric weighting function}
In order to quantify the contribution of each LiDAR residual during sensor fusion, we propose a metric weighting function according to the association quality as follows

\begin{equation*}
\srv_\circ(\vp^\cl)=\lambda\cdot(\bnu_\circ)^\top\vn_\circ\cdot\exp\big(\textstyle-\sum_{j=1}^5\vert\gamma(\vp^\cl)-\gamma_j\vert\,\big)\,.
\end{equation*}
Here, $\circ$ indicates the type of feature correspondences, namely edges (subscript `e') or planes (subscript `s'). For an edge feature correspondence, $\bnu_\Tte$ and $\vn_\Tte$ denote the direction vector of the edge line at $\vp^\cl$ and the line approximated by its nearest five edge features, respectively. Similarly, $\bnu_\Ts$ and $\vn_\Ts$ denote the plane normal at point $\vp^\cl$ and the one formed by its nearest five plane features, respectively. Moreover, $\gamma(\vp^\cl)$ and $\gamma_j$ are the reflectance values of the feature point $\vp^\cl$ and its associated nearest five features, respectively. Therefore, the proposed metric weighting function considers both geometric and appearance consistencies of feature associations. We set the constant $\lambda=15$ from experience.

\subsection{Feature-based scan-matching}
At the frontend, we perform a light-weight, feature-based scan-matching in a frame-to-model manner. A trust region method is used to estimate the current pose $\vx^\cw$ by minimizing
\begin{equation*}
\min_{\vx^\cw}\bigg\{{\sum_{i=1}^{m}\big(\calD_\Tte(\vx^\cw,\vp_i^\cl,\mM^\cw)\big)^2+\sum_{j=1}^{n}\big(\calD_\Ts(\vx^\cw,\vp_j^\cl,\mM^\cw)\big)^2}\bigg\}\,,
\end{equation*}
which incorporates all the edge ($\vp_i^\cl\in\mE$) and plane ($\vp_j^\cl\in\mF$) correspondences between the current scan and the local map $\mM^\cw$.

Here, $\calD_\Tte$ and $\calD_\Ts$ denote the point-to-edge and point-to-plane metrics, respectively. The frontend local feature map $\mM^\cw$ is updated with a width of typically $20$ recent frames given the optimized poses. The cost function above is similar to other popular LiDAR odometry systems~\cite{liosam2020shan,livoloam2019,zhang2014loam,ye2019tightly}. However, we restrict the optimization time for fast motion estimation, based on which the translational scan distortion is corrected and keyframes are selected. The accuracy of state estimation is pursued using the proposed backend fusion scheme shown below.

\section{Tightly-Coupled LiDAR-Inertial Fusion via Keyframe-Based Sliding Window Optimization} \label{sec:keyframeOpt}

\begin{figure}[t]
	\centering
	\includegraphics[width=0.95\textwidth]{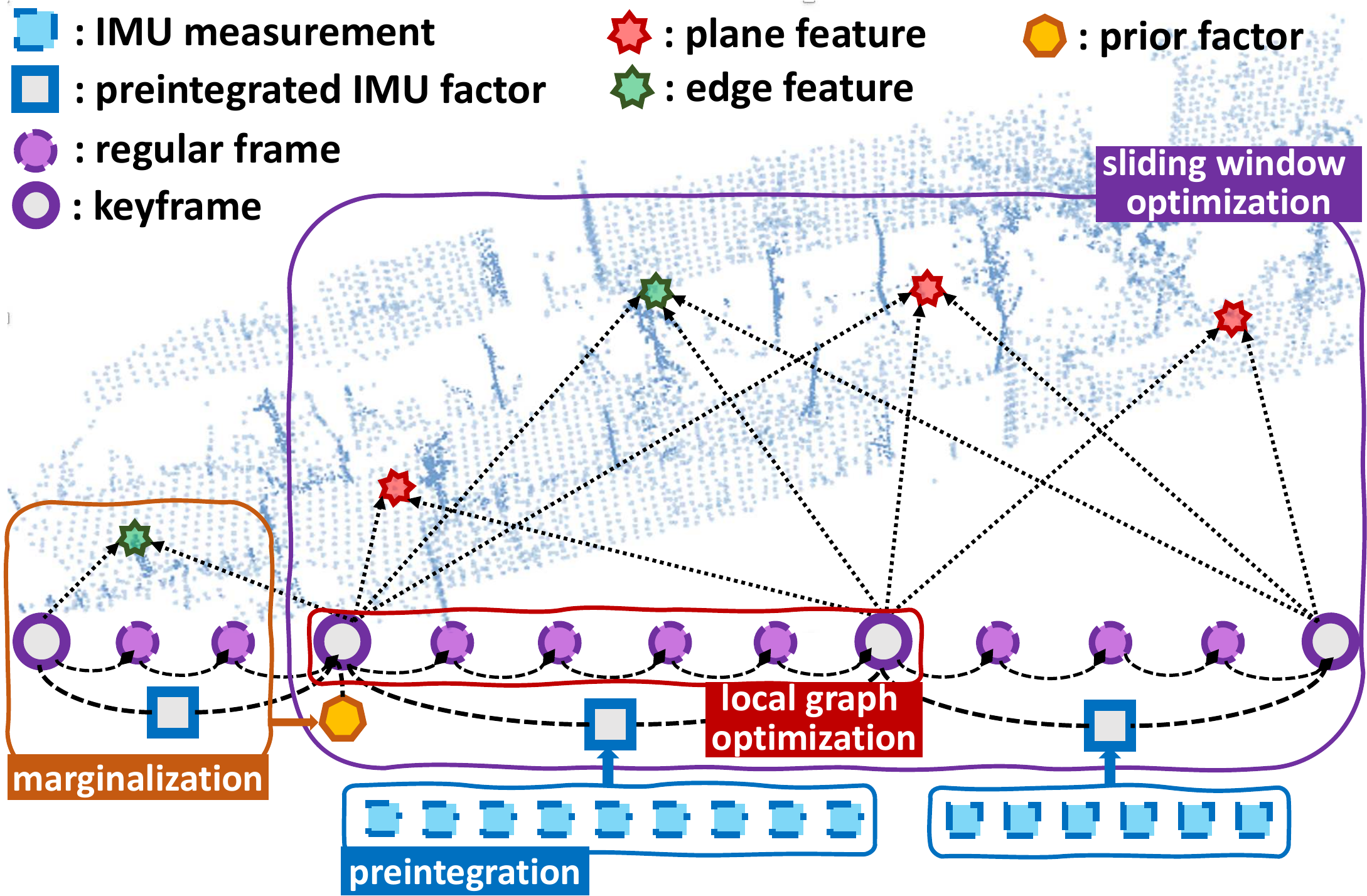} 
	\caption{Proposed tightly-coupled LiDAR-inertial fusion scheme using keyframe-based sliding window optimization.}	
	\label{fig:opt}
\end{figure}	

\subsection{Keyframe-based fusion hierarchy} \label{subsec:hie}
Keyframe-based schemes have originally been proposed and widely applied to visual odometry to achieve accurate tracking in real time~\cite{leutenegger2015keyframe}. In~\cite{liosam2020shan}, keyframes from LiDAR odometry frontend~\cite{legoloam2018shan} are fused to constrain IMU factors via iSAM2 (thus indirect fusion). The tightly-coupled LiDAR-inertial odometry in~\cite{ye2019tightly} realized direct fusion of LiDAR and pre-integrated IMU measurements via sliding window optimization. However, real-time performance is not generally achievable as the scheme fuses LiDAR sweeps of every frame. 

Therefore, it is of importance to maintain the sparsity of the optimization scheme for direct LiDAR-inertial fusion at the backend. Shown in Fig.\,\ref{fig:opt}, the proposed  fusion scheme exploits keyframes to establish sliding windows, where LiDAR and pre-integrated IMU measurements at keyframes are fused in a unified manner via nonlinear optimization. As the window slides forward after optimization, we construct a local factor graph incorporating the two oldest keyframe poses as constraints and the regular frames poses initialized by IMU measurements. A small-scale factor graph optimization is invoked to obtain regular-frame poses at LiDAR frequency.

Setting up keyframes can critically affect odometry accuracy due to the IMU drift during the time interval of two consecutive keyframes. We introduce two criteria for keyframe selection: (1)~If the overlapping ratio between features of the current frame and the local feature map is smaller than $60\%$ or (2)~if the time difference to the last keyframe is more than a certain number of (e.g., two) regular frames, the current frame is then selected as a new keyframe. Here, restricting the frame interval between keyframes helps mitigate the drift issue for IMU pre-integration.

\subsection{Sliding window optimization for keyframes} \label{subsec:opt}
We compute keyframe states of form~\eqref{eq:keyframe} by directly fusing LiDAR and pre-integrated IMU measurements, all observed at keyframes. For a sliding window of $\tau_\Tw$-keyframe width, the optimal keyframe states $\bfX=[\,\bvx_1^\top,....,\bvx_{\tau_\Tw}^\top\,]^\top\in\R^{15\times\tau_\Tw}$ are obtained by minimizing

\begin{equation} \label{eq:opt}
\min_{\bfX^\cw}\bigg\{\Vert\calR_\text{P}(\breve{\fX}^\cw)\Vert^2+\sum_{k=1}^{\tau_\Tw}\calJ_\text{L}(\bvx_k^\cw)+\sum_{k=1}^{\tau_\Tw}\calJ_\text{I}(\bvx_k^\cw)\bigg\}
\end{equation}
in the form of maximum a posterior (MAP). $\calR_\text{P}(\breve{\fX}^\cw)$ denotes the prior residual term representing the measurements that are marginalized out due to window-sliding. $\calJ_\text{L}(\bvx_k^\cw)$ and $\calJ_\text{I}(\bvx_k^\cw)$ denote the keyframe-wise LiDAR  and IMU error terms. Details about the three components follow.

\subsubsection*{Prior factor}
In order to bound the computational burden without substantial information loss, we exploit marginalization in the sliding window optimization. Here, the oldest keyframe and its measurements are marginalized out via Schur-complement~\cite{vins2018}. A new prior is computed accordingly and added on top of the existing prior factor to carry the estimate from the removed keyframe to the next window.

\subsubsection*{LiDAR term}
The LiDAR term incorporates geometric constraints from LiDAR measurements into the fusion scheme. When aligning the observed edge ($\bvp_{k,i}^\cl\in\breve{\mE}_k$) and plane ($\bvp_{k,j}^\cl\in\breve{\mF}_k$) features to the local feature map $\breve{\mM}_k^\cw$ observed by recent $30$ keyframes, the term is defined as $\calJ_\text{L}(\bvx_k^\cw)=$
\begin{equation*}
\smashoperator[lr]{\sum_{i=1}^{m}}\tilde{\srv}_{\Tte,i}\big(\calD_\Tte(\bvx_k^\cw,\bvp_{k,i}^\cl,\breve{\mM}_k^\cw)\big)^2+\smashoperator[lr]{\sum_{j=1}^{n}}\tilde{\srv}_{\Ts,j}\big(\calD_\Ts(\bvx_k^\cw,\bvp_{k,j}^\cl,\breve{\mM}_k^\cw)\big)^2.
\end{equation*}
Here, $\calD_\Tte$ and $\calD_\Ts$ are the point-to-edge and point-to-plane metrics in~Sec.\,\ref{subsec:lidarMetric} with fixed feature correspondences, respectively. $\tilde{\srv}_\Tte$ and $\tilde{\srv}_\Ts$ denote normalized weights among feature correspondences of each type. The local map is updated given the optimized poses as the window slides.

\subsubsection*{IMU term}
The error term for IMU incorporates the relative motion constraints between keyframes into the fusion scheme. To avoid repropagating IMU states each time the optimization window slides, raw inertial readings are pre-integrated between two consecutive keyframes $k$ and $k+1$ as in~\cite{vins2018,ye2019tightly}. The term is defined as
\begin{equation*}
\begin{aligned}
\calJ_\text{I}&(\bvx_k^\cw)=\Vert\calR_\text{I}(\bvx_k^\cw,\hat{\vz}^k_{k+1})\Vert_{\fC^k_{k+1}}^2,\text{with}\, \calR_\text{I}(\bvx_k^\cw,\hat{\vz}^k_{k+1})=\\
&\bbmat\,\breve{\fR}_k^{\cw\top}\big(\breve{\vt}^\cw_{k+1}-\breve{\vt}^\cw
_k+\frac{1}{2}\vg^\cw\Delta\tau_k^2-\vv^\cw_k\Delta\tau_k\big)-\hat{\balpha}^k_{k+1}
\\\breve{\fR}_k^{\cw\top}\big(\bvv^\cw_{k+1}+\vg^\cw{\Delta \tau_k}-\bvv^\cw_k\big)-\hat{\bbeta}^k_{k+1}
\\2\big[(\vq^\cw_k)^{-1}\otimes\vq^\cw_{k+1}\otimes(\hat{\bgamma}^k_{k+1})^{-1}\big]_\text{vec}
\\\bvb_{\Ta,k+1}-\bvb_{\Ta,k}
\\\bvb_{\Tg,k+1}-\bvb_{\Tg,k}
\,\ebmat
\end{aligned}
\end{equation*}
being the preintegrated measurement residual at keyframe $k$. $\hat{\vz}^k_{k+1}=[\,\hat{\balpha}^{k\top}_{k+1},\hat{\bbeta}^{k\top}_{k+1},\hat{\bgamma}^{k\top}_{k+1}\,]^\top$ is the pre-integrated IMU measurements incorporating gyroscope and accelerometer readings from keyframe $k$ to $k+1$. $\Delta\tau_k$ denotes the time interval between consecutive keyframes $k$ and $k+1$. We use the operator $[\,\cdot\,]_\text{vec}$ to take out the vector part of a quaternion. Due to space constraints, we do not provide the derivation of IMU pre-integration and the corresponding noise covariance $\fC^k_{k+1}$ for the Mahalanobis norm above. A dedicated introduction can be found in~\cite{vins2018}. 

The nonlinear least square problem in~\eqref{eq:opt} can be solved using typical solvers, e.g., the trust region method. Constrained by the optimized keyframes, regular LiDAR frames in between are obtained via local graph optimization. After the fusion window slides over, the newly obtained LiDAR poses are inserted into a global pose graph with only keyframe feature maps maintained for mapping purpose. To detect potential loop closures, we search in the global graph, e.g., in a radius of $10$\,m, to find keyframe nodes that are spatially close but with enough temporal distance (e.g., $20$ keyframes). An ICP is performed between the current feature scan and the candidate feature map, from which a fitting score is computed for loop closure detection. Once confirmed, a global pose graph optimization is invoked by imposing the constraint from the ICP. As depicted in Fig.\,\ref{fig:flow}, the keyframe local map $\breve{\mM}^\cw$ at backend is then updated by corrected poses to further incorporate the LiDAR constraint.

\section{Evaluation} \label{sec:eva}
\subsection{Implementation and evaluation setup}	\label{subsec:implement}
We implement the proposed LiDAR-inertial odometry and mapping system in C++ using ROS~\cite{quigley2009ros}. The three modules shown in Fig.\,\ref{fig:flow} are structured as three individual nodes. The nonlinear optimization problem in~\eqref{eq:opt} is solved using the Ceres Solver~\cite{ceres}. We use GTSAM~\cite{dellaert2012factor} to perform factor graph optimization for rectifying the global pose graph at loop closures. Our system is developed for Livox Horizon with the name LiLi-OM. It is, however, also applicable for conventional spinning LiDARs thanks to its generic backend fusion. Thus, two versions of the system are evaluated: (1)~the original LiLi-OM for Livox Horizon with the proposed feature extraction approach and (2)~its variant $\listar$ using the preprocessing module of~\cite{zhang2014loam} for spinning LiDARs. Evaluations are conducted based on public data sets (recorded using conventional LiDARs) and experiments (including data sets from Livox Horizon). All LiDAR frame rates are $10$ Hz.

\subsection{Public data set} \label{subsec:dataSet}
We deploy $\listar$ to compare with competing state-of-the-art systems. These include works on (1) LiDAR odometry: A-LOAM\footnote{\tt https://github.com/HKUST-Aerial-Robotics/A-LOAM} (open-source version of LOAM~\cite{zhang2014loam}), LeGO-LOAM~\cite{legoloam2018shan} (shortened as LeGO) and (2) LiDAR-inertial odometry: LIO-mapping (shortened as LIOM)~\cite{ye2019tightly}, LINS~\cite{lins2020}, LIO-SAM~\cite{liosam2020shan}. For evaluation, we use the EU long-term data set (\textit{UTBM}) that provides two long urban navigation sequences recorded by a Velodyne HDL-32E and a six-axis IMU ($100$ Hz)~\cite{yan2019eu}. LIO-SAM requires nine-axis IMU measurements. Thus, we include the \textit{UrbanLoco} and \textit{UrbanNav} data sets~\cite{wen2020urbanloco} recorded using a HDL-32E and Xsens MTi-10 IMU (nine-axis, $100$ Hz). The RMSE of the absolute position error (APE) is computed for the final estimated trajectory based on the ground truth using the script in~\cite{grupp2017evo}. 

Shown in Tab.\,\ref{tab:comp}\footnote{\scriptsize Data sets abbr.: \textit{UTBM-1}: UTBM-20180719, \textit{UTBM-2}: UTBM-20180418-RA, \textit{UL-1}: UrbanLoco-HK-20190426-1, \textit{UL-2}: UrbanLoco-HK-20190426-2, \textit{UN-1}: UrbanNav-HK-20190314, \textit{UN-2}: UrbanNav-HK-20190428.}, the proposed $\listar$ achieves the best tracking accuracy (bold) for all sequences in real time. LIO-SAM~\cite{liosam2020shan} requires nine-axis IMU readings for de-skewing and frontend odometry, thereby not applicable for \textit{UTBM}. For the remaining sequences, LIO-SAM still shows worse tracking accuracy than LiLi-OM though it additionally exploits orientation measurements from a magnetometer. This mainly results from the unified fusion scheme of LiLi-OM where LiDAR and inertial measurements are directly fused. LIOM fails on \textit{UrbanNav} data sets (denoted as \xmark) and shows large drift on \textit{UTBM-1}. It also cannot run in real time with recommended configurations. LOAM delivers large tracking errors on \textit{UTBM} as the implementation limits the iteration number in scan-matching for real-time performance.

\begin{table}[H]
	\centering
	\caption{APE (RMSE) in meters on public data sets}
	\begin{tabular}{|c|c|c|c|c|c|c|}
		\hline
		dataset & LOAM &  LeGO & LIOM & LINS & LIO-SAM & $\listar$\\
		\hline
		\textit{UTBM-1}  & $479.51$&$17.12$ &$468.75$&$16.90$& -- &$\textbf{8.61}$\\
		\textit{UTBM-2}  & $819.95$&$6.46$&$12.95$&$9.31$& -- &$\textbf{6.45}$\\
		\textit{UL-1}    &$2.39$&$2.22$&$2.53$&$2.27$&$2.54$& $\textbf{1.59}$\\
		\textit{UL-2}    &$2.58$&$2.30$&$2.00$&$2.99$&$2.50$& $\textbf{1.20}$\\
		\textit{UN-1}    &$11.20$&$2.70$& \xmark &$2.19$&$2.28$&$\textbf{1.08}$\\
		\textit{UN-2}    &$12.70$&$4.15$& \xmark &$4.80$&$5.31$&$\textbf{3.24}$\\
		\hline
	\end{tabular}
	\label{tab:comp}
\end{table}

\subsection{Experiment} \label{subsec:exp}
To further test LiLi-OM in real-world scenarios, we set up a sensor suite composed of a Livox Horizon and an Xsens MTi-670 IMU. The total cost is about $1700$ Euros (Q1, 2020), which is much less than conventional LiDAR-inertial setups. 
\begin{figure}[t]
	\centering
	\begin{tabular}{cc}
		\includegraphics[width=0.62\textwidth]{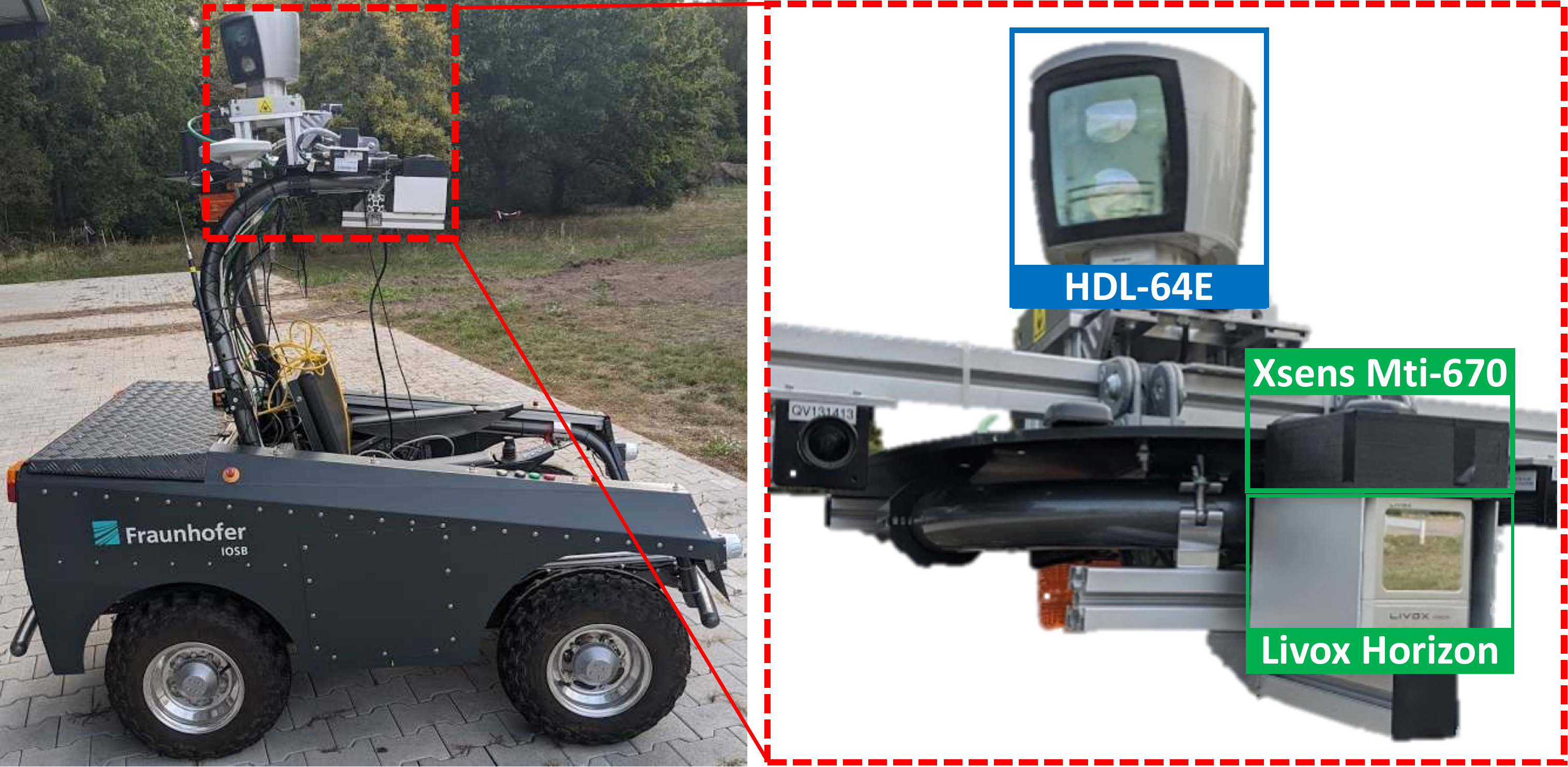} &	
		\includegraphics[width=0.34\textwidth]{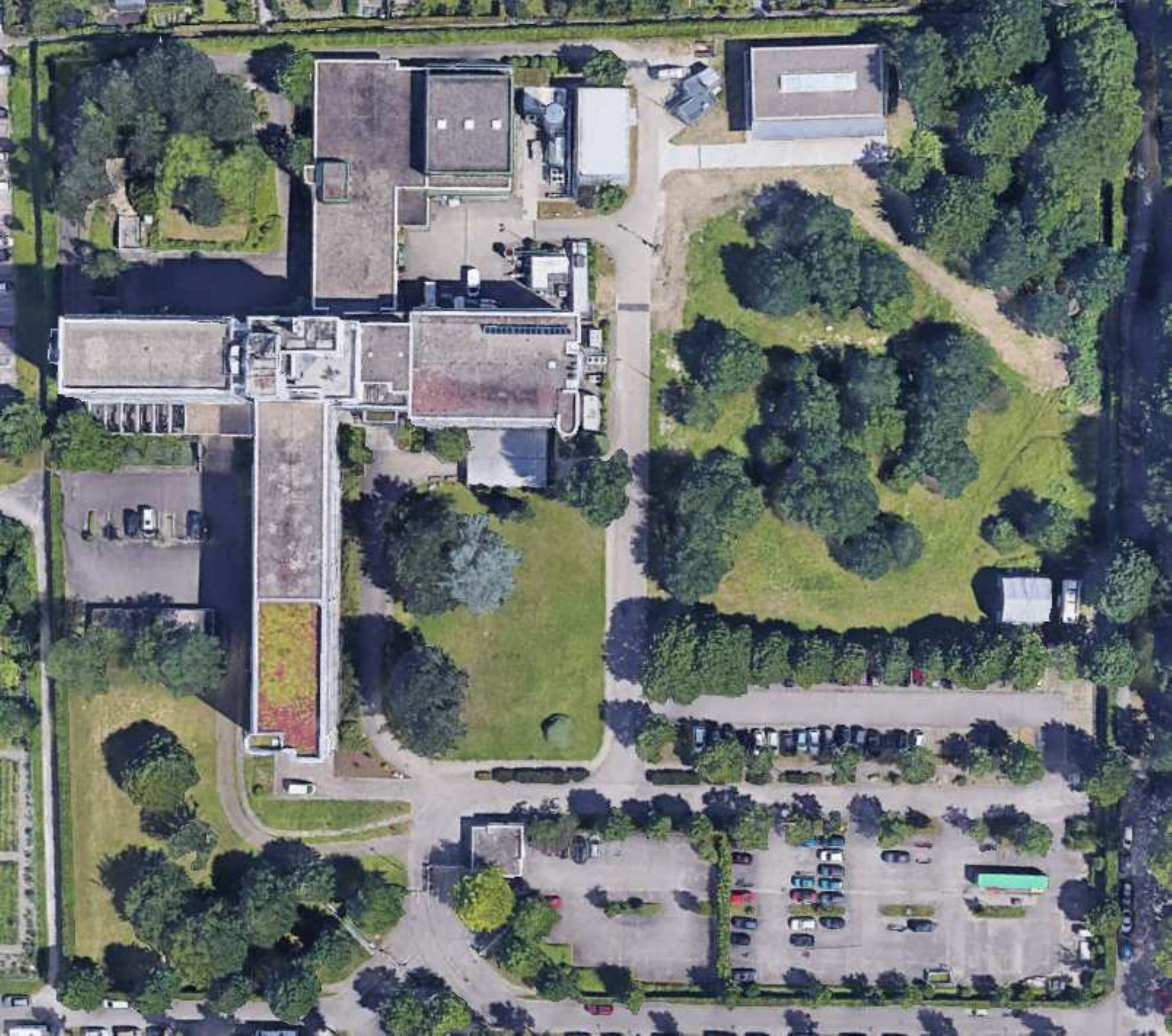} \\
		(A) & (B)
	\end{tabular}
	\caption{Experimental setup for \textit{FR-IOSB} data set.}
	\label{fig:car}
\end{figure}

\subsubsection{FR-IOSB data set} 
Shown in Fig.\,\ref{fig:car}-(A), a mobile platform is instrumented with the proposed Livox-Xsens suite. For comparison with high-end mechanical spinning LiDARs, we set up a Velodyne HDL-64E onboard and synchronize it with an Xsens MTi-G-700 IMU (six-axis, $150$ Hz). Three sequences were recorded at the Fraunhofer IOSB campus of Fig.\,\ref{fig:car}-(B): (1) \textit{Short} for a short path in structured scenes, (2) \textit{Tree} recorded in bushes, and (3) \textit{Long} for a long trajectory.  

\begin{figure}[t]
	\vspace{2mm}
	\centering
	\begin{tabular}{cc}
		\includegraphics[width=0.441\textwidth]{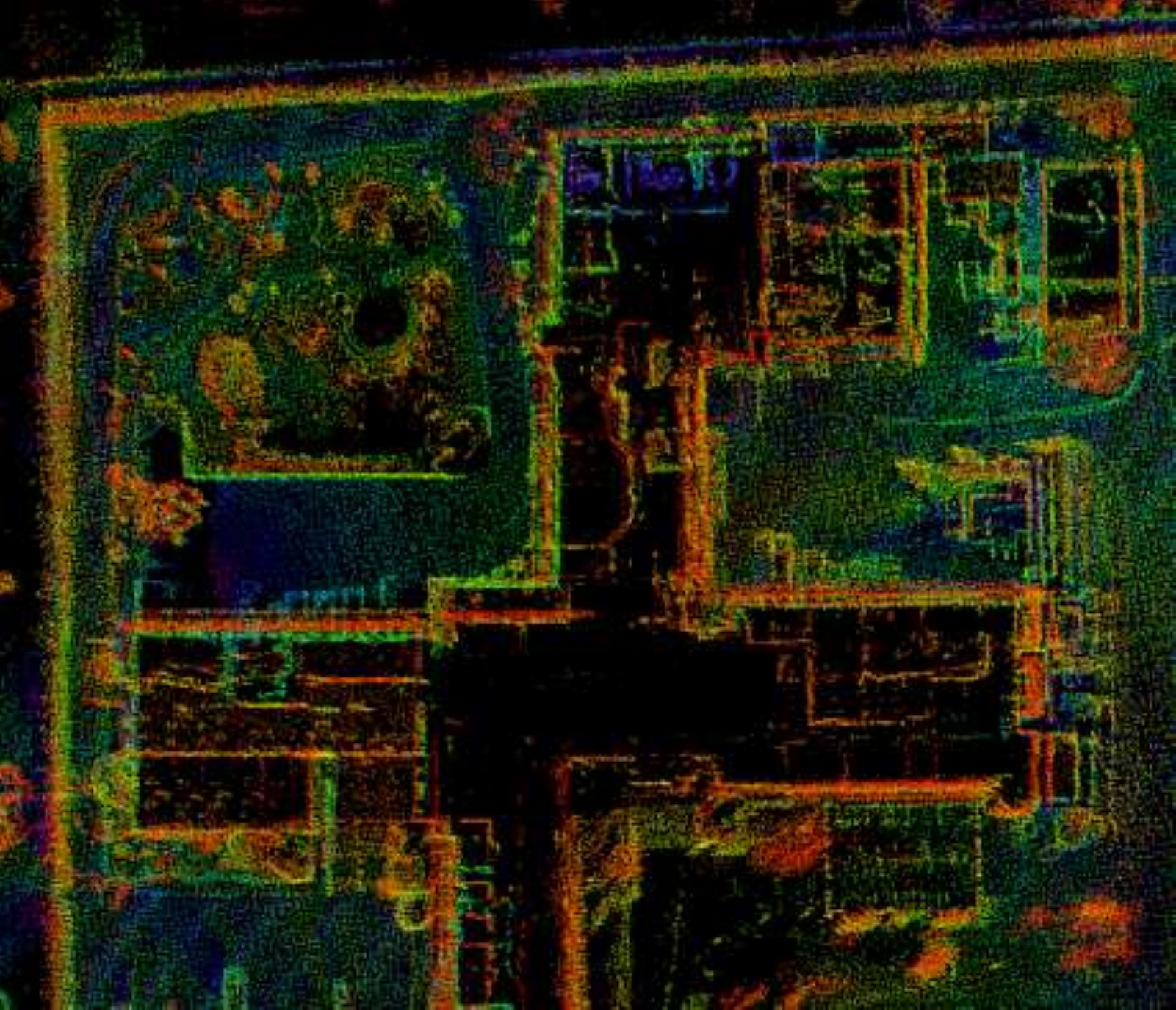} &
		\includegraphics[width=0.45\textwidth]{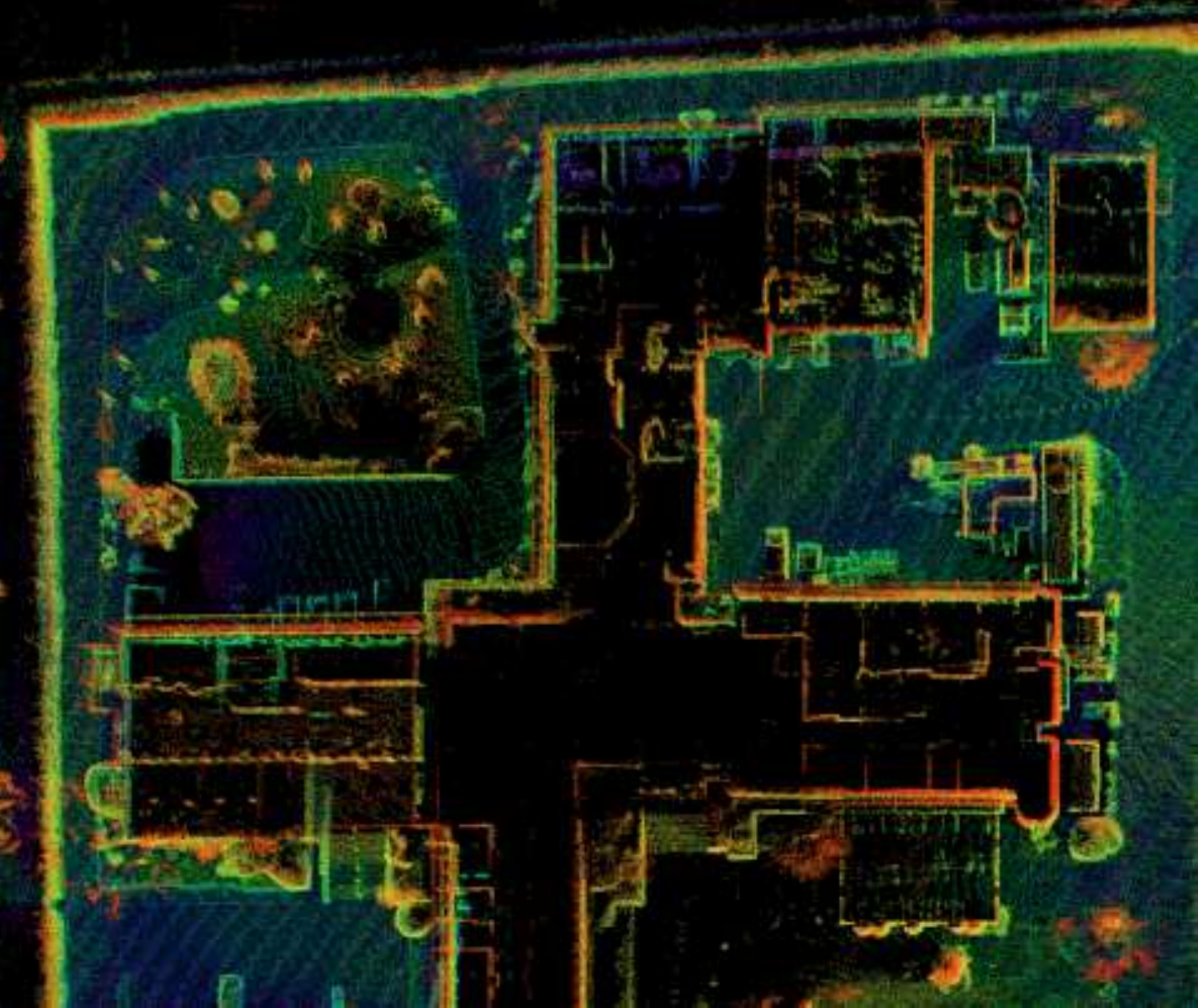} \\
		(A) LOAM & (B) $\listar$
	\end{tabular}
	\caption{Mapping result comparison on sequence \textit{Long}.}
	\label{fig:iosb}
\end{figure}

Both LiLi-OM and $\listar$ are tested. For comparison, we run LOAM, LeGO and Livox-Horizon-LOAM (shortened as LiHo)\footnote{\tt https://github.com/Livox-SDK/livox\_horizon\_loam}, a LOAM variant adapted to Livox Horizon with point clouds deskewed  by IMU. Tab.\,\ref{tab:iosb} shows superior tracking accuracy (bold) of the proposed systems with our low-cost hardware setup performing equally well as the high-end one in the same scenario. We show reconstructed maps (partial) on sequence \textit{Long} in Fig.\,\ref{fig:iosb}, where $\listar$ delivers superior mapping quality using the proposed sensor fusion scheme.

\begin{table}[H]
	\centering
	\caption{End-to-end position error in meters on \textit{FR-IOSB}}
	\begin{tabular}{|c|c|c|c|c|c|c|c|}
		\hline
		\multicolumn{3}{|c|}{}  & \multicolumn{3}{c|}{Velodyne HDL-64E} & \multicolumn{2}{c|}{Livox Horizon} \\ 
		\hline
		dataset& length & speed & LOAM & LeGO & $\listar$ & LiHo & LiLi-OM  \\
		\hline
		\textit{Short} & $0.49$~km & $2.15$~m/s & $0.78$ & $\bf0.25$ & $\bf0.34$ & $5.04$ & $\bf0.25$\\
		\textit{Tree}  & $0.36$~km & $1.12$~m/s & $0.21$ & $78.22$ & $\bf{<0.1}$ & $0.13$ & $\bf{<0.1}$
		\\
		\textit{Long}  & $1.10$~km & $1.71$~m/s & $0.43$ & $0.82$ & $\bf{<0.1}$ & $3.91$ & $\bf0.34$\\
		\hline
	\end{tabular}
	\label{tab:iosb}
\end{table}

\subsubsection{KA-Urban data set} 
\begin{figure}[t]
	\centering
	\begin{tabular}{cc}
		\includegraphics[width=0.22\textwidth]{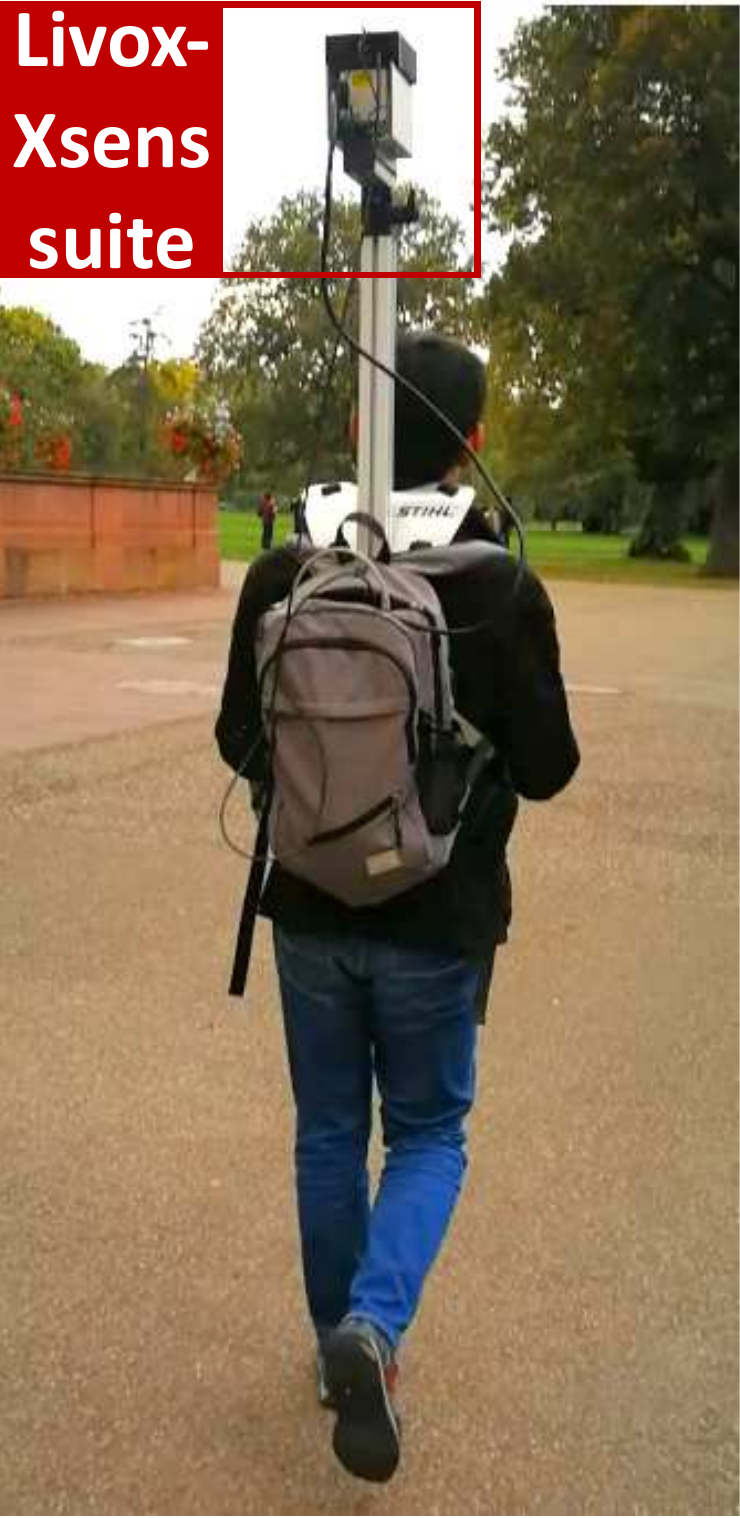} &
		\includegraphics[width=0.7\textwidth]{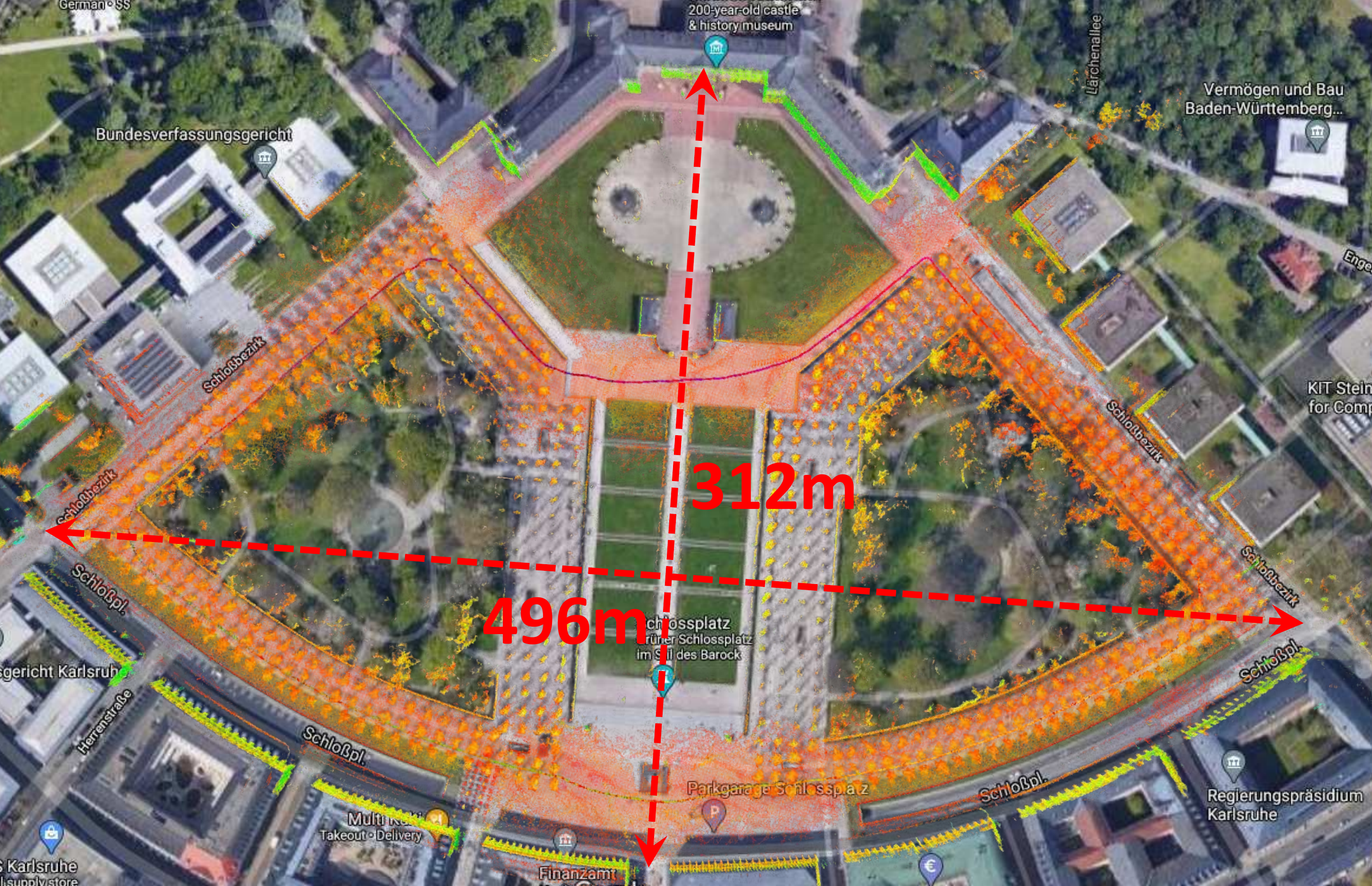} \\
		(A) &(B) 
	\end{tabular}
	\caption{(A) Recording \textit{KA-Urban} with proposed sensor suite. (B) Map from LiLi-OM on \textit{Schloss-2} (path ends at origin).}
	\label{fig:bg}
\end{figure}

Shown in Fig.\,\ref{fig:bg}-(A), the proposed Livox-Xsens sensor suite was further deployed onboard a backpack platform for large-scale test in urban scenarios. Five sequences were recorded in Karlsruhe, Germany with end-to-end locations registered from satellite images. Beside standard configuration of LiLi-OM, we deactivate its loop closure module to evaluate the odometry accuracy without global correction. LiHo is run for comparison. 

Shown in Tab.\,\ref{tab:urban}, LiLi-OM without loop closure (denoted as LiLi-OM-O) delivers much less drift than LiHo. When exploiting loop closure constraints, LiLi-OM shows very small end-to-end errors. In order to justify the benefits of multi-sensor fusion, we provide another special configuration of LiLi-OM with IMU constraints totally removed (including de-skewing). Evaluations are done on long-distance sequences (the last three ones) with loop closure both on and off. Denoted by the second entry after ``\,/\,", this configuration is superior to LiHo (also pure LiDAR odometry), but inferior to the standard one using LiDAR-inertial fusion.

\begin{figure}[t]
	\begin{tabular}{cc}
		\includegraphics[width=0.45\textwidth]{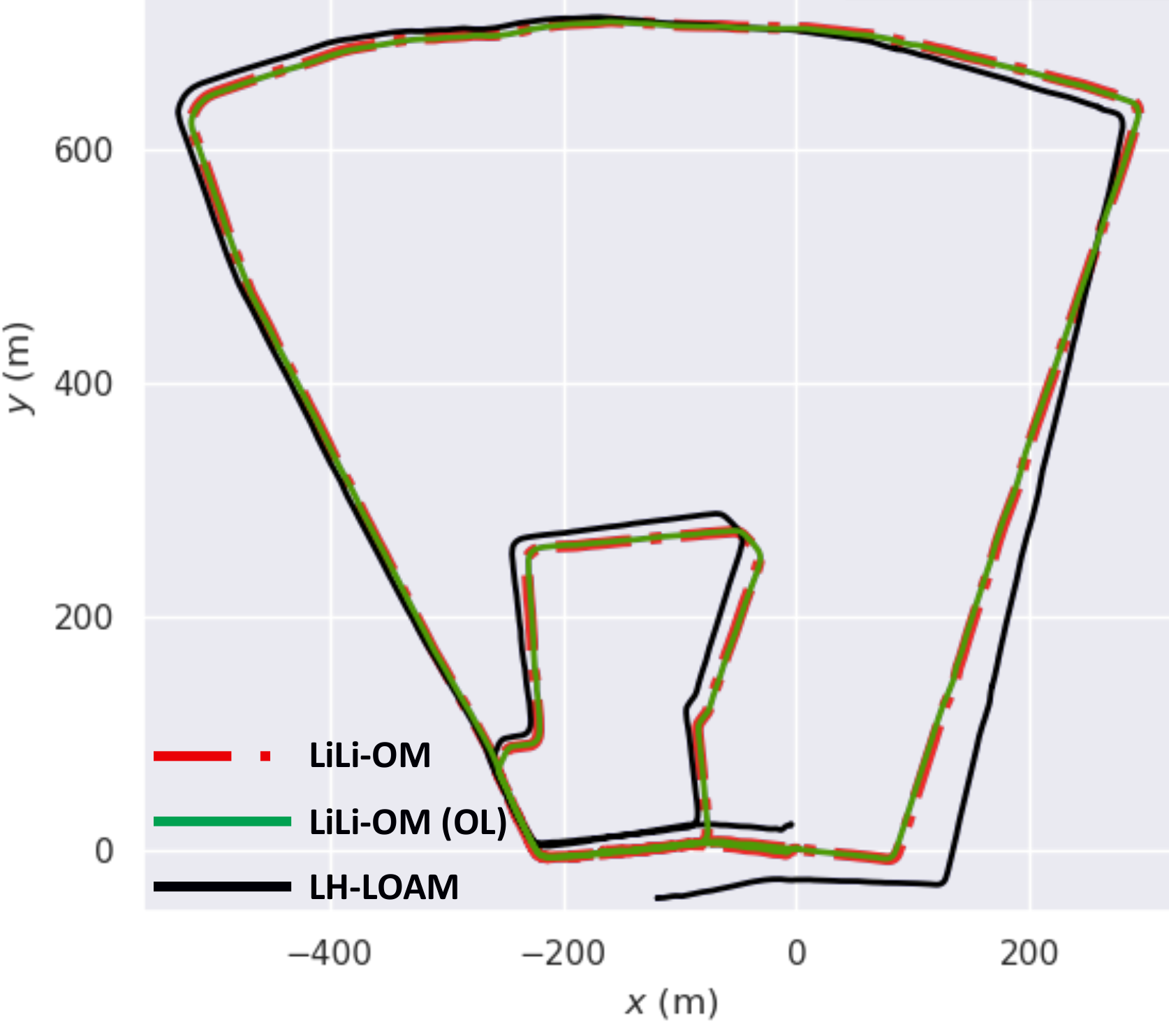} &
		\includegraphics[width=0.42\textwidth]{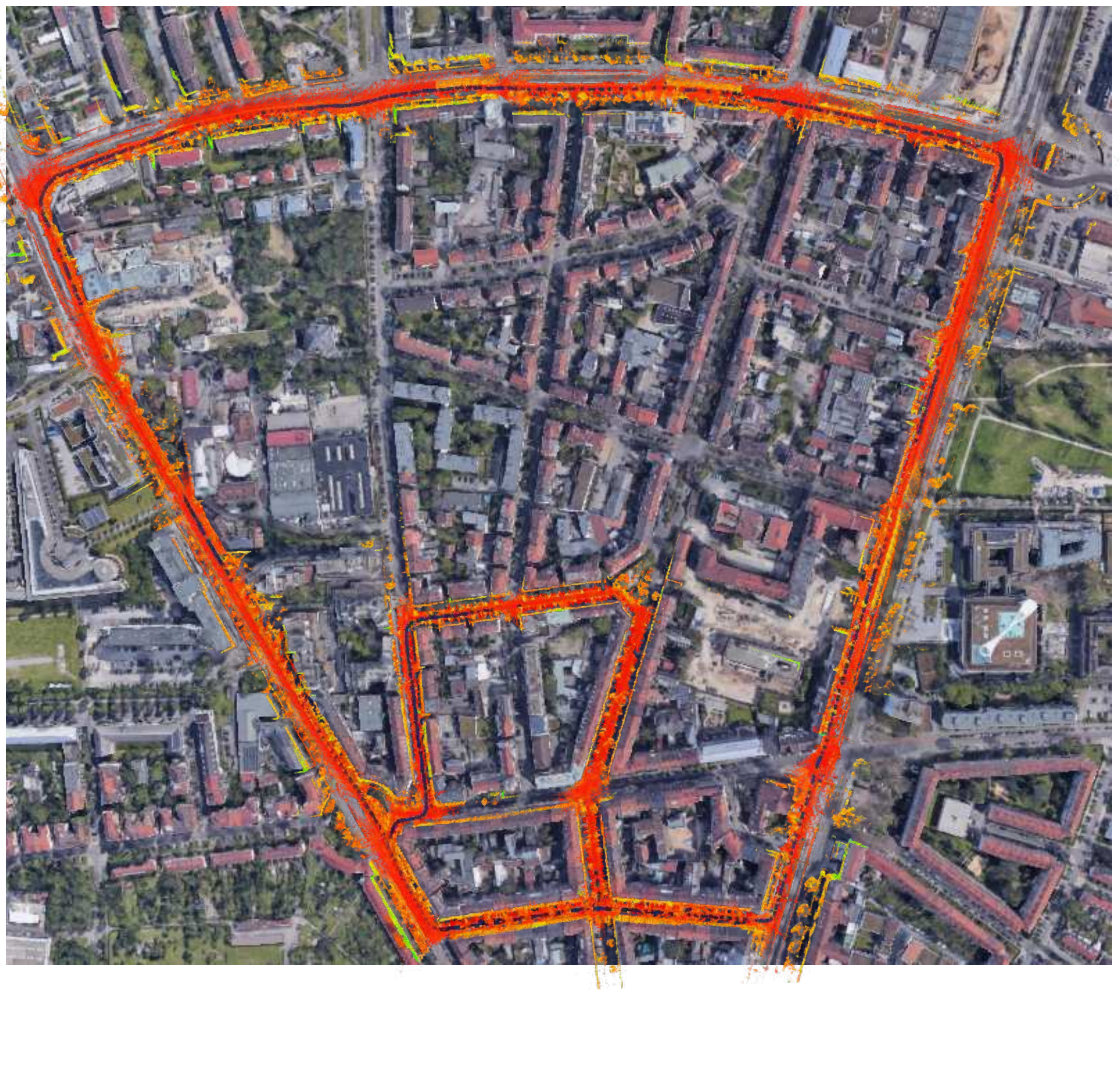} \\
	\end{tabular}
	\caption{Test results on sequence \textit{East}.}
	\label{fig:urban}
\end{figure}

The mapping result of LiLi-OM on \textit{Schloss-1} is given in Fig.\,\ref{fig:front}, where the Schloss Karlsruhe is digitalized in high-precision point cloud using the proposed sensor suite. Result of running LiLi-OM on \textit{Schloss-2} is visualized in Fig.\,\ref{fig:bg}-(B). An area of $496\,\text{m}\times312\,\text{m}$ is mapped accurately with global consistency (compared to satellite image). Sequence \textit{East} was recorded while cycling through eastern Karlsruhe for a long distance under fast and dynamic motion. Shown in Fig.\,\ref{fig:urban}, LiLi-OM delivers accurate odometry and mapping results using the proposed low-cost sensor suite. 

\begin{table}[H]
	\centering
	\caption{End-to-end position error in meters on \textit{KA-Urban}}
	\begin{tabular}{|c|c|c|c|c|c|}
		\hline
		dataset & length & speed & LiHo &  LiLi-OM-O & LiLi-OM  \\
		\hline
		\textit{Campus-1}  & $0.50$\,km & $1.43$\,m/s& $1.47$ & $1.11$\,/\,--& $\bf{0.13}$\,/\,--\\
		\textit{Campus-2}  & $0.20$\,km & $1.57$\,m/s& $0.40$ & $0.21$\,/\,--& $\bf{0.19}$\,/\,--\\
		\textit{Schloss-1} & $0.65$\,km & $1.03$\,m/s& $1.55$ & $0.95$\,/\,$1.21$& $\bf{0.15}$\,/\,$0.24$\\
		\textit{Schloss-2} & $1.10$~km & $1.49$~m/s& $8.34$& $4.41$\,/\,$5.58$& $\bf{0.08}$\,/\,$0.12$\\
		\textit{East} & $3.70$~km & $3.11$~m/s& $109.62$& $15.66$\,/\,$19.28$ & $\bf{1.28}$\,/\,$3.43$\\
		\hline
	\end{tabular}
	\label{tab:urban}
\end{table}

\subsection{Runtime} \label{subsec:runtime}
All evaluations are done on a laptop (Intel Core i5-7300HQ CPU, 8GB RAM) with all four CPU cores involved. For all the data sets recorded with different devices in the evaluation, LiLi-OM delivers real-time performance (at LiDAR frame rate).  The three nodes in Fig.\,\ref{fig:flow} run in parallel and their average runtime per frame on representative sequences is collected in Tab.\,\ref{tab:runtime}. Preprocessing for feature extraction and scan registration at frontend are light-weight. Runtime is dominated by backend fusion.

\begin{table}[H]
	\centering
	\caption{Runtime of LiLi-OM per frame in ms}
	\begin{tabular}{|c|c|c|c|c|c|c|}
		\hline
		& \multicolumn{3}{c|}{Velodyne HDL} & \multicolumn{3}{c|}{Livox Horizon} \\ 
		\hline
		node & \textit{UTBM-2}&\textit{UL-1}&\textit{Long} &\textit{Long}&\textit{Schloss-2}&\textit{East} \\
		\hline
		preprocessing &$12.72$&$13.31$&$30.14$&$9.99$&$10.21$&$11.76$ \\
		scan registration &$22 .29$&$23.71$&$16.71$&$22.69$&$27.15$&$25.30$ \\
		backend fusion &$50.27$&$57.62$&$60.92$&$58.86$&$54.56$&$41.81$ \\
		\hline
	\end{tabular}
	\label{tab:runtime}
\end{table}

Constructing the LiDAR constraints at backend fusion often takes substantial time. In Tab.\,\ref{tab:feature}, we show statistics on feature extraction for LiLi-OM with typical configurations. Extracted features are counted per frame at preprocessing, while only associated features are counted at backend fusion per keyframe. Due to the relatively uniform FoV coverage of Livox Horizon, more plane features are extracted than edges on average. To guarantee runtime efficiency, features are always downsampled using a voxel grid filter before providing the LiDAR constraints.

\begin{table}[H]
	\centering
	\caption{Average feature numbers}
	\begin{tabular}{|c|c|c|c|c|c|}
		\hline
		& \multicolumn{3}{c|}{preprocessing /\,frame} & \multicolumn{2}{c|}{backend fusion /\,keyframe} \\ 
		\hline
		dataset & raw points & edges   & planes & ~~~~edges~~~~ & planes  \\
		\hline
		\textit{Schloss-1}  & $22579$ & $207$ & $12791$ & $162$ & $3688$\\
		\textit{Schloss-2}  & $22230$ & $633$ & $13128$ & $425$ & $2649$
		\\
		\textit{East}  & $21544$ & $337$ & $13109$ & $247$ & $2363$\\
		\hline
	\end{tabular}
	\label{tab:feature}
\end{table}

\section{Conclusion} \label{sec:conc}
In this work, we propose a novel sensor fusion method for real-time LiDAR-inertial odometry and mapping. A keyframe-based hierarchical scheme is established for directly fusing LiDAR and (pre-integrated) IMU measurements via sliding window optimization. Given the optimized keyframe states, regular-frame poses are obtained via factor graph optimization. The proposed LiDAR-inertial odometry and mapping system is universally applicable for both conventional LiDARs and solid-state LiDARs of small FoVs. For the latter use case, a novel feature extraction method is designed for the irregular and unique scan pattern of Livox Horizon, a newly released spinning free, solid-state LiDAR with much lower price than conventional 3D LiDARs. We conduct evaluations on both public data sets of conventional LiDARs and experiments using the Livox Horizon. Results show that the proposed system is real-time capable and delivers superior tracking and mapping accuracy over state-of-the-art LiDAR/LiDAR-inertial odometry systems. The proposed system, LiLi-OM, is featured as a cost-effective solution of high-performance LiDAR-inertial odometry and mapping using solid-state LiDAR. 

There is still much potential to exploit for the proposed system. The deployed Livox-Xsens suite is lightweight and LiLi-OM is developed for universal egomotion estimation (not only for planar motion as~\cite{legoloam2018shan}). Thus, it should be, for instance, tested onboard unmanned aerial vehicles in applications such as autonomous earth observation or environmental modeling coping with aggressive six-DoF egomotion. For large-scale odometry and mapping with limited computational resources, advanced map representations can be employed to improve memory as well as runtime efficiency. Potential options include volumetric mapping using TSDF (Truncated Signed Distance Fields)~\cite{voxgraph2020} or mapping with geometric primitives (especially in man-made environment~\cite{ICRA20_Li}). 

\section*{Acknowledgment}
We would like to thank Thomas Emter from Fraunhofer IOSB for providing the robot platform and assistance in recording the \textit{FR-IOSB} data set. 

	\bibliographystyle{IEEEtran.bst}
	\bibliography{./bibliography.bib}
	
\end{document}